%% file: neurips_2024.tex
\documentclass{article}

\usepackage[preprint]{neurips_2024}

\usepackage{xspace}
\usepackage{multirow}
\usepackage{multicol}
\usepackage{amsmath}
\usepackage{amssymb}
\usepackage{mathtools}
\usepackage{amsthm}
\usepackage[pdftex]{graphicx}
\usepackage{subfig}
\usepackage{floatrow}
\usepackage{wrapfig}
\usepackage{caption}
\usepackage{natbib}
\newfloatcommand{capbtabbox}{table}[][\FBwidth]
\newtheorem{theorem}{Theorem}

\usepackage[utf8]{inputenc} 
\usepackage[T1]{fontenc}    
\usepackage{hyperref}       
\usepackage{url}            
\usepackage{booktabs}       
\usepackage{amsfonts}       
\usepackage{nicefrac}       
\usepackage{microtype}      
\usepackage{xcolor}         

\newcommand{\gbdt}{\textsc{GBDT}\xspace}
\newcommand{\gbdts}{\textsc{GBDTs}\xspace}
\newcommand{\pfn}{\textsc{PFN}\xspace}
\newcommand{\pfns}{\textsc{PFNs}\xspace}
\newcommand{\tabpfn}{\textsc{TabPFN}\xspace}
\newcommand{\tabpfnc}{\textsc{TabPFN*}\xspace}
\newcommand{\icl}{\textsc{ICL}\xspace}
\newcommand{\icp}{\textsc{ICP}\xspace}
\newcommand{\icps}{\textsc{ICPs}\xspace}
\newcommand{\micp}{\textsc{MICP}\xspace}
\newcommand{\capfn}{\textsc{CaPFN}\xspace}
\newcommand{\ours}{\textsc{MixturePFN}\xspace}
\newcommand{\tz}{\textsc{TabZilla}\xspace}

\newcommand{\xgb}{\textsc{XGBoost}\xspace}

\title{Mixture of In-Context Prompters for Tabular PFNs}

\author{%
    Derek Xu\textsuperscript{1}, 
    Olcay Cirit\textsuperscript{2}, 
    Reza Asadi\textsuperscript{2}, 
    Yizhou Sun\textsuperscript{1}, 
    Wei Wang\textsuperscript{1}
    \\ \\
    \textsuperscript{1}University of California, Los Angeles \\ \textsuperscript{2}Uber AI
}

\begin{document}

\maketitle

\input{sec-abstract.tex}
\input{sec-introduction.tex}

\input{sec-method.tex}

\input{sec-experiments.tex}
\input{sec-conclusion.tex}

\bibliographystyle{plainnat}
\bibliography{bibliography}
\input{sec-appendix.tex}

\end{document}

%% file: sec-abstract.tex
\begin{abstract}

Recent benchmarks found In-Context Learning (ICL) outperforms both deep learning and tree-based algorithms on small tabular datasets. However, on larger datasets, ICL for tabular learning cannot run without severely compromising performance, due to its quadratic space and time complexity w.r.t. dataset size. We propose \ours, which both extends nearest-neighbor sampling to the state-of-the-art ICL for tabular learning model and uses bootstrapping to finetune said model on the inference-time dataset. \ours is the Condorcet winner across 36 diverse tabular datasets against 19 strong deep learning and tree-based baselines, achieving the highest mean rank among Top-10 aforementioned algorithms with statistical significance.

\end{abstract}

%% file: sec-introduction.tex
\section{Introduction}

Tabular data is a popular data format across various domains, consisting of column-wise features and row-wise data samples. Each feature can be either continuous, categorical, or ordinal. Thanks to the prevalence of relational databases, which ensure data integrity, consistency, and low redundancy, tabular data is widely used across various domains such as medicine, finance, and advertising. Hence, improving learning algorithms on tabular data is of interest to many researchers.

General tabular datasets remain unconquered by most deep learning algorithms~\citep{popov2019neural, gorishniy2021revisiting, somepalli2021saint,arik2021tabnet,yamada2020feature,yoon2020vime, chen2022danets}. Instead, gradient-boosted decision trees (\gbdts)~\citep{chen2016xgboost, prokhorenkova2018catboost}, achieve better overall performance on tabular benchmarks \citep{mcelfresh2023neural, shwartz2022tabular} considering a wide range of number of samples, numbers of features, feature types, and feature distributions. Recently, transformer-based prior-fitted networks, \pfns~\citep{hollmann2022tabpfn}, have garnered interest, for their surprisingly strong and state-of-the-art performance on tabular datasets with $\le 3,000$ samples~\citep{hollmann2022tabpfn, mcelfresh2023neural}.

 Unlike SGD-based deep learning, \pfns learn the training algorithm itself~\citep{muller2021transformers,von2023transformers}. Specifically, a \pfn first pretrains a model by sampling labelled datasets from a predefined dataset prior~\citep{muller2021transformers}, then performs inferece using In-Context Learning (\icl)~\citep{brown2020language}, where each downstream dataset is tokenized into a ``prompt'', sent to the \pfn model, which returns test split predictions, as described in Section~\ref{sec:pfn}. By learning the learning algorithm itself, \pfns discover better inductive bias than conventional deep learning algorithms and gradient-boosted decision trees, which require extensive hyperparameter tuning to effectively fit the downstream dataset~\citep{hollmann2022tabpfn}.

\pfns cannot scale to datasets with $>3000$ samples without compromising performance due to two key limitations. First, \pfn inference is computationally expensive. Because the entire dataset is fed to the transformer as a ``prompt'', the inference time and space complexity for $N_{train}$ training samples is $\mathcal{O}(N_{train}^2)$. This is in stark contrast to \gbdts and traditional deep learning approaches where the inference time and space complexity is $\mathcal{O}(1)$.  To fit larger datasets in memory, existing works~\citep{hollmann2022tabpfn, mcelfresh2023neural} resort to randomly sampling the training dataset during inference when $N_{train}>3000$ to ensure the ``prompt'' has at most 3000 samples. Second, existing \pfns assume the dataset prior used during pretraining is always representative of the dataset prior used during inference, which we empirically find is untrue. Specifically, downstream performance improves when the \pfn is finetuned on how datasets are sampled during inference rather than during pretraining.

In this work, we analyze recent claims~\citep{mcelfresh2023neural} on \pfn's effectiveness, finding it does not scale w.r.t. dataset size. We improve \pfn's scalability by proposing Sparse ``Mixture of In-Context Prompters'' (\micp), which creates scalable ``prompts'' by routing new test samples to a group of prompters. We solve \pfn's alignment limitations with ``Context-Aware \pfn'' (\capfn), which finetunes \pfns for downstream datasets via bootstrapping. We call our combined model \ours. To summarize:

\begin{itemize}

  \item To improve scalability, we are the first to propose Sparse Mixture of In-Context Prompters (\micp) which routes new test samples to a pool of scalable prompters for In-Context Learning. \micp reduces \pfn inference complexity from $\mathcal{O}(N_{train}^2)$ memory and time to $\mathcal{O}(1)$ memory and $\mathcal{O}(log(N_{train}))$ time, w.r.t. the number of training samples, $N_{train}$.
  
  \item To improve performance, we finetune Context-Aware Prior-Fitted Network (\capfn), which aligns pretrained \pfns with inference-time datasets using a novel bootstrapping policy.
  
  \item \ours scales transformer \pfns from tabular datasets of 3,000 samples to those with much larger number of samples, with no performance deterioration w.r.t. dataset size. 
  
  \item \ours is the Condorcet winner across 36 diverse tabular datasets against 19 strong deep learning and tree-based baselines, achieving the top mean rank among Top-10 aforementioned algorithms with statistical significance.

\end{itemize}

%% file: sec-method.tex
\section{{Preliminaries}}

We consider tabular classification problems, where the inputs are numerical, ordinal, or categorical columns encoded as a $d$-dimensional feature vector, $x\in\mathbb{R}^{d}$, the output is the corresponding label, $y\in[1,...,C]$, and the dataset consists of labelled input output pairs, $D = \{(x^{(i)}, y^{(i)})\}_{i=0}^{N}$.\footnote{We provide a table with all math notations in the Supplementary Material.}  Given the training dataset, $D_{train} = \{(x^{(i)}_{train}, y^{(i)}_{train})\}_{i=0}^{N_{train}}$, and test samples, $X_{test} = [x_{test}^{(i)}]_{i=0}^{N_{test}}$,
our goal is to correctly predict the corresponding test labels, $Y_{test} = [y_{test}^{(i)}]_{i=0}^{N_{test}}$. \ours is inspired by Prior Fitted Networks, which we first introduce.

\subsection{Prior Fitted Networks}
\label{sec:pfn}

Prior Fitted Network (\pfn)~\citep{muller2021transformers} is a parameterized model, $q_{\theta}$, that learns to approximate Bayesian inference given the dataset prior, $p(D)$, via In-Context Learning (\icl)~\citep{brown2020language}. Specifically, \pfn inference approximates the posterior predictive distribution (PPD), $p_\theta(y|x,D) = \int_\phi p(y|x,\phi) p(D|\phi) p(\phi) d\phi$, where $\phi$ is the hypothesis mechanism behind how the tabular data is generated. For example, $\phi$ can be a structural causal model. Refer to Muller et. al~\citep{muller2021transformers} for further details.

\begin{figure*}
  \centering
  \includegraphics[width=0.9\textwidth]{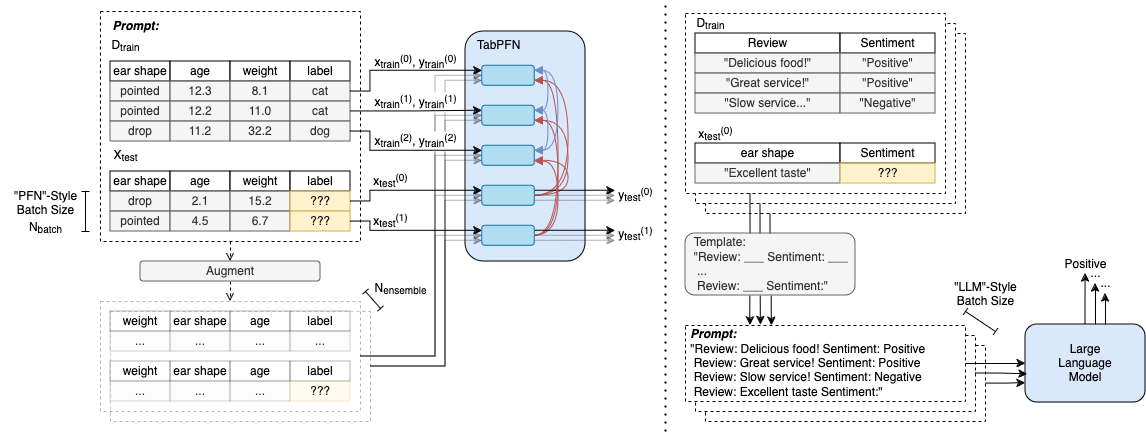} 
  \caption{We highlight the differences between In-Context Learning (\icl) on Prior Fitted Networks (ex. \tabpfn), left, and  Large Language Models (LLMs), right. \tabpfn treats training data as tokens (where each token is a concatenation of feature and label), whereas LLMs use templates to convert training data into natural language prompts. \tabpfn uses an attention pattern (blue and red arrows) supporting batch inference, whereas LLMs use generic encoder-decoder or decoder-only setups. \tabpfn are pretrained on Equation~\ref{eqn:pfn_loss}, whereas LLMs are pretrained on a separate objective.}
  \label{fig:icl}
\end{figure*}

\subsubsection{Pretraining}

To approximate the PPD, \pfns are pretrained to minimize KL-Divergence between the parameterized model, $q_{\theta}(y|x,D)$, and the PPD, $p(y|x,D)$, over the dataset prior, $p(D)$, which was proven equivalent to optimizing the prior data negative log likelihood, $\mathcal{L}_{\pfn}$. As shown in Equation~\ref{eqn:pfn_loss}~\citep{muller2021transformers}, this loss iteratively samples new datasets from a handcrafted dataset prior, $p(D)$, via Monte-Carlo. 

\begin{equation}
\label{eqn:pfn_loss}
\mathcal{L}_{\pfn} = \mathop{\mathbb{E}}_{\substack{x,y,D\sim p(D)}} [-log(q_\theta(y|x,D))]
\end{equation}

\tabpfn~\citep{hollmann2022tabpfn} is the state-of-the-art pretrained \pfn transformer for tabular data. It treats the hypotheses, $\phi$, as randomly sampled structural causal models (SCM)~\citep{pearl2009causality,peters2017elements} mixed with the original Bayesian Neural Network prior~\citep{muller2021transformers}. Training dataset samples are generated by first sampling a SCM graph, $\phi \sim p(\phi)$, followed by sampling the SCM, $x,y,D \sim p(D|\phi)$. 

Transformer-based~\citep{vaswani2017attention} \pfns tokenize the sampled dataset, $(x,D)$ as input to the parameterized model, $q_\theta$, as shown in Figure~\ref{fig:icl} and discussed in Section~\ref{sec:inf}. Note, \pfn inputs are analogous to ``prompts'' from In-Context Learning (ICL)~\citep{brown2020language, dong2022survey, xu2024context}, hence they are called ``prompts'' in this work. 

\subsubsection{Inference}
\label{sec:inf}

During inference, transformer-based \pfns tokenize the downstream dataset, $(X_{test},D_{train})$, into batched ``prompts'', consisting of $N_{train}$ encoder tokens and $N_{batch}$ decoder tokens, where each data sample corresponds with one token.\footnote{Our dataset is split into train/dev/test sets. During hyperparameter tuning, decoder tokens are taken from the dev set instead.}  Because tabular columns are permutation invariant, \tabpfn shuffles feature orderings and scalings, running $q_\theta$ on each permutation of the ``prompt'', then returning an ensembled prediction, as depicted in Figure~\ref{fig:icl} and further detailed in Section~\ref{subsec:batch}. \tabpfn does not perform finetuning, only inference, on downstream datasets.

\subsubsection{Fundamental Scalability Limitations}

\tabpfn cannot scale to datasets with large numbers of training samples, $N_{train} \geq 3,000$~\citep{mcelfresh2023neural}. Because the ``prompt'' contains $N_{train}$ encoder tokens, the transformer's, $q_\theta$, inference space and time complexities are quadratic w.r.t. dataset size, $\mathcal{O}(N_{train}^2)$, which is too computationally expensive to run on large datasets.\footnote{While linear transformer approximations are $\mathcal{O}(N_{train})$, our goal is to scale \pfns to be comparable to other predictive algorithms which have $\mathcal{O}(1)$ space and time complexity w.r.t. the number of training samples.} Conventional deep learning~\citep{popov2019neural,arik2021tabnet,gorishniy2021revisiting} and \gbdt~\citep{chen2016xgboost, prokhorenkova2018catboost} algorithms have $\mathcal{O}(1)$ inference space and time complexities w.r.t. number of training samples. Our proposal, \ours, reduces the space complexity of \tabpfn inference to $\mathcal{O}(1)$ and time complexity to $\mathcal{O}(log(N_{train}))$, effectively scaling \pfns to larger datasets.

\subsubsection{\pfn-Style Batching}
\label{subsec:batch}

Batching in \tabpfn is unlike in Large Language Models (LLMs). LLMs for In-Context Learning (ICL) can fit $N_{batch}$ test samples across $N_{batch}$ ``prompts''~\citep{liu2021makes}, where each test sample has its own ``prompt''. \tabpfn~\citep{hollmann2022tabpfn} must fit $N_{batch}$ test samples in $1$ ``prompt'', because the ensembling process will augment each ``prompt'' into $N_{ensemble}$ inputs through shuffling feature orderings and scalings. We focus on \textbf{\tabpfn-style batching which fits $N_{batch}$ test samples in $1$ ``prompt''}.

\begin{figure*}
  \centering
  \includegraphics[width=0.9\textwidth]{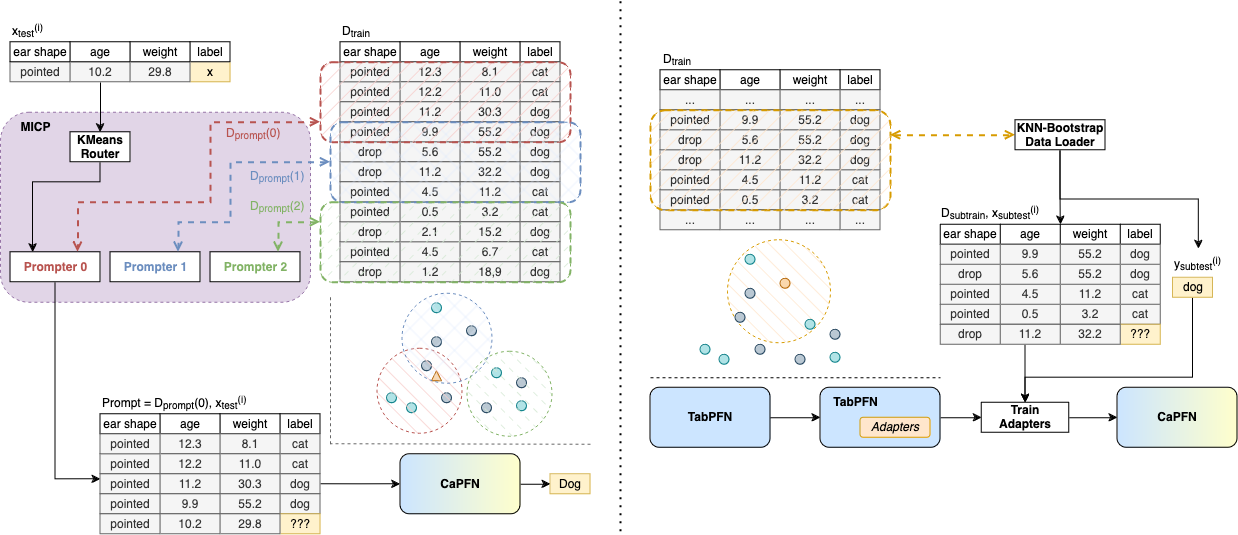} 
  
  \caption{Illustration of \ours. \textbf{\micp (Left)}: New test samples are passed to a router that picks 1 out of $K$ prompters to form a scalable ``prompt'' with $B$ training samples for the downstream \pfn model. \textbf{\capfn (Right)}: \tabpfn is frozen, fitted with adapters, then finetuned using data prior negative loss likelihood, Equation~\ref{eqn:pfn_loss}, on our bootstrapped data prior, $p(D|D_{train})$. This prior simulates the \micp inference mechanism. The finetuned model is called \capfn.}
  
  \label{fig:micp_capfn}
\end{figure*}

\section{Method}

\subsection{Support Set Approximation for Scalable \pfns}
\label{sec:approx}

To improve space and time complexity, \ours hypothesizes each test sample, $x_{test}^{(i)}$, requires only a small support set, $D_{supp}(x_{test}^{(i)}) \subset D_{train}$, of constant size, $|D_{supp}(x_{test}^{(i)})|=B$ $\forall i=[0,...,N_{test}-1]$, to effectively perform inference with In-Context Learning. This assumption is reasonable as training samples with drastically different features from the test sample should have little impact on its label~\citep{khandelwal2019generalization,liu2021makes,xu2023k,feuer2023scaling}. For example, the purchase trends of today share more in common with purchase trends from last week than the those from 10 years ago. 

We define the support set, $D_{supp}$, for each individual test sample, $x_{test}^{(i)}$, to be the $B$ spatially closest training samples, which could be found via K-Nearest Neighbors: $D_{supp}(x) = \textsc{KNN}(x| D_{train}, B)$. A scalable \pfn ``prompt'' can thus be constructed with the test sample and support set, $(\{x_{test}^{(i)}\}, D_{supp}(x_{test}^{(i)}))$. We call this approach KNN-Prompting~\footnote{Further illustrations of KNN-Prompting and its relation to ICL for LLMs can be found in the Appendix.}.


\pfn inference on KNN prompts, $(\{x_{test}^{(i)}\}, D_{supp}(x_{test}^{(i)}))$, can be computed in $\mathcal{O}(1)$ space and time complexity w.r.t. $N_{train}$ for fixed $B$. However, each test sample, $x_{test}^{(i)}$, may require a different support set, $D_{supp}(x_{test}^{(i)})$, and hence its own ``prompt''. Thus, KNN-Prompting does not support \tabpfn-style batching, where multiple test cases fit in 1 ``prompt''. Furthermore, KNN-Prompting runs an expensive KNN search across the whole training dataset for each test sample. For these 2 reasons, KNN-Prompting is too expensive in practice when evaluating on larger tabular datasets.

\subsection{Mixture of In-Context Prompters (\micp)}
\label{sec:micp}

In order to efficiently construct effective ``prompts'', \ours leverages cluster structures in the data. Inspired by Sparse Mixture of Experts~\citep{shazeer2017outrageously,lewis2021base}, where each test sample is routed to an specialized expert trained on a subset of the training dataset, Sparse Mixture of In-Context Prompters, \micp, routes each test sample to one of $K$ ``In-Context Prompters'' (\icp), $\{\mathcal{T}_k\}_{k=0}^K$, specializing on a cluster of the training dataset, using a routing module, $\mathcal{R}: \mathbb{R}^d \rightarrow \{0, ..., K-1\}$. Each \icp then constructs a relevant ``prompt'' for incoming test samples, which are sent to the downstream \pfn model in batch.

To reduce the ``prompt''-construction overhead, each \icp, $\{\mathcal{T}_k\}_{k=0}^K$, precomputes its own support set, $D_{prompt}(k) \subset D_{train}$, of constant size, $|D_{prompt}(k)|=B$. During inference, \icp concatenates incoming test samples with its support set to form the scalable ``prompt'': $\mathcal{T}_k(\{x_{test}^{(i)}:\mathcal{R}(x_{test}^{(i)}=k)\}) = (\{x_{test}^{(i)}:\mathcal{R}(x_{test}^{(i)}=k)\}, D_{prompt}(k))$. The goal of \micp is efficiently approximate KNN-prompts, which can be accomplished by maximizing the overlap between the prompter's and the test sample's support sets: $|D_{prompt}(\mathcal{R}(x_{test}^{(i)})) \cap D_{supp}(x_{test}^{(i)})|$ $\forall i \in [0, ..., N_{test}-1]$.

\subsubsection{Router and Prompter Initialization}

Intuitively, each \icp, $\mathcal{T}_k$, specializes on a local cluster of the training dataset. The router, $\mathcal{R}$, routes each test sample to its nearest cluster.  To capture local clusters, we initialize the router and \icp with K-Means on the training dataset: $\{D_{cluster}^{(k)}\}_{k=0}^{K}, \{x_{center}^{(k)}\}_{k=0}^{K} = \textsc{KMeans}(D_{train})$. 
Our resulting router is the Nearest Neighbor Search algorithm: $\mathcal{R}(x) = \textsc{NNS}(x | \{x_{center}^{(k)}\}_{k=0}^{K})$.

Because efficient ``prompts'' have bounded number of entries, $B$, we subsample clusters with more than $B$ entries. Because \pfn performance drastically increases with more training samples~\citep{hollmann2022tabpfn, muller2021transformers, mcelfresh2023neural}, we expand clusters with less than $B$ entries via K-Nearest Neighbors. We define this process in Equation~\ref{eqn:icp}, where  $G(k) = |D_{cluster}^{(k)}| < B$  denotes whether the clusters are too big or small.

\begin{equation}
\label{eqn:icp}
D_{prompt}(k) = \\
\begin{cases}
    \textsc{KNN}(x_{center}^{(k)}| D_{train}, B), & \text{if } G(k) \\
    \textsc{Sample}(D_{cluster}^{(k)}, B),              & \text{else}
\end{cases}
\end{equation}

During inference, \micp routes each test sample, $x_{test}^{(i)}$, to its corresponding \icp. Similar to sparse mixture of experts, once each \icp receives enough test entries, \pfn inference is performed on \textbf{batched} test ``prompts'': $(X_{batch},D_{prompt}^{(k)})$, which contains multiple entries from the test set $X_{batch} \subseteq X_{test}$, because all said entries were routed to the same \icp cluster: $\mathcal{R}(x_{batch}^{(i)}) = \mathcal{R}(x_{batch}^{(j)}) \forall i,j$. Hence, \micp efficiently constructs bounded batched ``prompts''.

\begin{table*}
\centering
\begin{tabular}{l|llll|ll}
\toprule
\multirow{ 2}{*}{\textbf{Method}} & \multicolumn{4}{c|}{\textbf{Condorcet Statistics}} & \textbf{All Algo.} & \textbf{Top-10 Algo.} \\
& \#Votes$\uparrow$ & \#Wins$\uparrow$ & \#Ties & \#Losses$\downarrow$ & \multicolumn{2}{c}{Mean $\pm$ Std Rank$\downarrow$}  \\
\midrule
MixturePFN & \textbf{524} & \textbf{19} & 0 & \textbf{0} & \textbf{2.350 $\pm$ 1.824} & \textbf{2.273 $\pm$ 1.7106} \\
XGBoost & 500 & 18 & 0 & 1 & 5.500 $\pm$ 4.621 & 4.000 $\pm$ 2.663 \\
CatBoost & 474 & 17 & 0 & 2 & 4.900 $\pm$ 4.158 & 3.955 $\pm$ 2.688 \\
SAINT & 408 & 16 & 0 & 3 & 8.300 $\pm$ 5.367 & 4.045 $\pm$ 1.965 \\
TabPFN* & 381 & 13 & 1 & 5 & 4.550 $\pm$ 2.747 & 4.040 $\pm$ 1.311 \\
LightGBM & 373 & 14 & 1 & 4 & 9.150 $\pm$ 4.351 & 6.409 $\pm$ 2.839 \\
DANet & 312 & 14 & 0 & 5 & 9.050 $\pm$ 3.369 & 7.045 $\pm$ 1.988 \\
FTTransformer & 294 & 12 & 0 & 7 & 8.600 $\pm$ 3.541 & 6.773 $\pm$ 2.235 \\
ResNet & 286 & 11 & 0 & 8 & 8.400 $\pm$ 3.262 & 6.864 $\pm$ 1.961 \\
SVM & 285 & 9 & 0 & 10 & 11.300 $\pm$ 4.766 & 7.500 $\pm$ 2.482 \\
STG & 284 & 10 & 0 & 9 & 11.900 $\pm$ 4.549 & - \\
RandomForest & 247 & 7 & 0 & 12 & 11.600 $\pm$ 4.443 & - \\
NODE & 243 & 7 & 0 & 12 & 13.350 $\pm$ 3.410 & - \\
MLP-rtdl & 227 & 5 & 0 & 14 & 10.800 $\pm$ 5.046 & - \\
TabNet & 210 & 5 & 0 & 14 & 13.550 $\pm$ 5.296 & - \\
LinearModel & 202 & 3 & 1 & 15 & 12.400 $\pm$ 4.652 & - \\
MLP & 191 & 5 & 1 & 13 & 13.700 $\pm$ 3.621 & - \\
VIME & 134 & 2 & 0 & 17 & 15.350 $\pm$ 3.851 & - \\
DecisionTree & 114 & 1 & 0 & 18 & 16.800 $\pm$ 3.881 & - \\
KNN & 74 & 0 & 0 & 19 & 18.450 $\pm$ 1.936 & - \\
\bottomrule
\end{tabular}
\caption{\ours is the Condorcet winner across 36 datasets against 19 baseline algorithms. \ours achieves the top mean rank across 20 datasets where all algorithms successfully run and across 22 datasets where all Top-10 algorithms successfully run. To break ties, we rank algorithms based on their mean log-likelihoods following \tz~\citep{mcelfresh2023neural}. We report the Condorcet matrix, dataset breakdowns, and accuracy-metric results in the Appendix.}
\label{tab:main}
\end{table*}

In total, router and prompter initialization takes $\mathcal{O}(tN_{train}K + (N_{train}+KB)logN_{train})$ time and $\mathcal{O}(N_{train} + KB)$ space complexity and is done once before inference. Routing takes $\mathcal{O}(log(K))$ time and $\mathcal{O}(1)$ space complexity, using efficient nearest neighbor search with ball-tree for each test sample. \pfn transformer inference takes $\mathcal{O}(B^2 + BN_{batch})$ time and space complexity, as \micp prompts contain at most $B$ training samples and $N_{batch}=|X_{batch}|$ testing samples. We provide time and space complexity details in the Appendix. We illustrate \micp in Figure~\ref{fig:micp_capfn}.

\subsubsection{Efficiency and Effectiveness Trade-Off}
\label{sec:trade}

The effectiveness of \micp prompts depend on the number of \icps used, $K$. As the complexity and size of data increase, more \icps are needed to capture the entropy of the labels. This is natural as each router's support set, $D_{prompt}(\mathcal{R}(x_{test}^{(i)}))$, should be representative of test samples routed to that cluster, $D_{support}(x_{test}^{(i)})$. If the true support set $D_{support}(x_{test}^{(i)})$ becomes more granular as the dataset size increases, more \icps are required to maximize overlap: $|D_{prompt}(\mathcal{R}(x_{test}^{(i)})) \cap D_{supp}(x_{test}^{(i)})|$.

We theoretically characterize this relationship between $K$, $B$, and overlap by analyzing conditions required for nonzero overlap on the training data: $|D_{prompt}(\mathcal{R}(x_{train}^{(i)})) \cap D_{supp}(x_{train}^{(i)})| \ge 1$ $\forall i \in [0, ..., N_{train}-1]$. Specifically, we encourage nonzero overlap by scaling the number of ``prompts'', $K$, linearly with the size of each ``prompt'', $B$, and training dataset size, $|N_{train}|$, as stated in Theorem~\ref{th:overlap1}: $K \ge \lceil N_{train}/B \rceil$.

This insight allows \ours to trade-off efficiency and effectiveness with a single hyperparameter, $\gamma$, which controls the number of \icps as a ratio of training and support set sizes: $K = \lceil \gamma N_{train}/B \rceil $. Intuitively, larger $\gamma$ improves effectiveness at the cost of efficiency. \textbf{Assuming fixed $\gamma$, $N_{batch}$, and $B$, \ours routing takes $\mathcal{O}(log(N_{train}))$ time and $\mathcal{O}(1)$ space complexity, and \pfn inference takes $\mathcal{O}(1)$ time and space complexity.}

\begin{theorem}[Nonzero Overlap]
\label{th:overlap1}
If every K-Means cluster contains at most $B$ samples, $|D_{cluster}^{(k)}| \leq B$ $\forall k\in[0,...,K-1]$ and training points route to their assigned K-Means cluster $\mathcal{R}^*(x_{train}^{(i)}) = k:x_{train}^{(i)} \in D_{cluster}^{(k)}$~\footnote{Thse conditions can be satisfied via constrained K-Means~\citep{bradley2000constrained}, which ensures each cluster has at most $B$ entries, and a router that sends train points to their assigned clusters. In practice, we find the relationship with the tunable parameter $\gamma$ also holds for \ours's regular K-Means and Nearest-Neighbor Search router.}, then nonzero overlap on the training data is guaranteed, $|D_{prompt}(\mathcal{R}^*(x_{train}^{(i)})) \cap D_{supp}(x_{train}^{(i)})| \ge 1$ $\forall i \in [0, ..., N_{train}-1]$ $\forall D_{train}$.
\end{theorem}

\begin{figure}
  \centering

  
  \subfloat[$N_{train}$ w.r.t. \tabpfnc\label{fig:scaletabpfn}]{\includegraphics[width=0.275\textwidth]{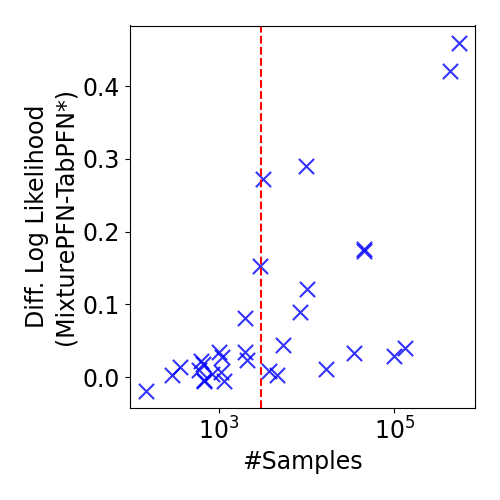}} 
  \subfloat[$N_{train}$ w.r.t. baselines\label{fig:scaleall}]{\includegraphics[width=0.275\textwidth]{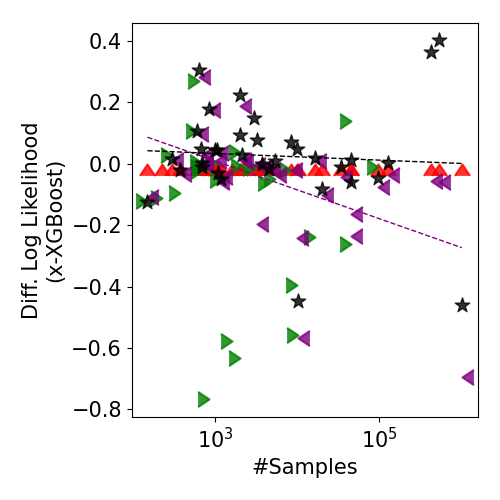}} 
  \subfloat[Kurtosis and \#Feats\label{fig:scaleprop}]{\includegraphics[width=0.45\textwidth]{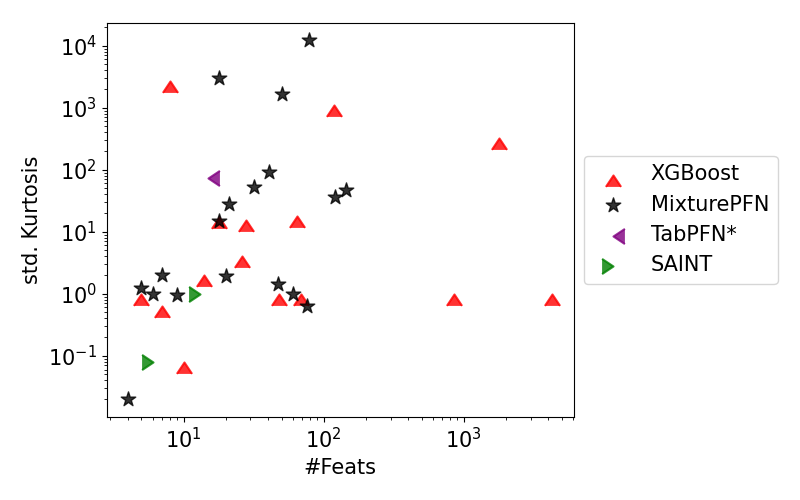}} 
  


  \caption{\textbf{(a):} We plot the difference in Log Likelihood between \ours and \tabpfnc for each dataset of size $N_{train}$. \ours substantially improves the performance and \tabpfnc and runs on datasets with $>3,000$ samples. \textbf{(b):} We plot the Log Likelihood of the top deep learning (DL) PFN, and tree baselines across all 36 datasets and the best-fit line between rank and dataset size, compared to the top baseline. Unlike \tabpfn, \ours maintains its good performance as the size of the dataset increases. \textbf{(c) :} We plot the best among the top DL, PFN, and tree baselines on all 36 datasets across different dataset properties. \ours performs well across different dataset irregularities. We provide further breakdowns in the Appendix.}
  \label{fig:kurtosis}
\end{figure}

\subsection{Context-Aware Finetuning (\capfn)}

\pfns are pretrained on the \icl task over a synthetic dataset prior, $p(D)=p(D|\phi)p(\phi)$~\citep{muller2021transformers,hollmann2022tabpfn}. Inspired by recent works which aligns Large Language Models on ICL ``prompts'' via finetuning~\citep{thoppilan2022lamda,wei2021finetuned, gu2023pre}, we argue the pretraining data prior, $p(D) = p(D|\phi)p(\phi)$, is different than the true data generating mechanism during inference, $p(D_{prompt}|D_{train})$, which was described in Section~\ref{sec:micp}. To better align the parameterized model, $q_{\theta}$, with the inference-time dataset, $D_{prompt}$, \capfn uses bootstrapping on the downstream dataset, $D_{train}$, to simulate ICL ``prompts'': $(X_{subtest}, Y_{subtest}, D_{subtrain}) \sim p(D|D_{train})$, where $X_{subtest} \subset X_{train}$, $Y_{subtest} \subset Y_{train}$, and $D_{subtrain} \subset D_{train}$. Bootstrapped samples are used to tune adapters~\citep{houlsby2019parameter} via prior data negative log likelihood loss, as shown in Equation~\ref{eqn:pfn_loss}, except the dataset prior is now the bootstrap mechanism: $p(D) = p(D|D_{train})$.

\subsubsection{Bootstrapping Large \micp Datasets}

The bootstrap procedure mimics \micp on large $N_{train} > 3000$ datasets: $p(D|D_{train}) = p(D_{support}|x)p(x|D_{train})$. Specifically, we sample a random training point from the training dataset, $x\sim p(x|D_{train})$, then run K-Nearest Neighbors from the sampled point, $p(D_{support}|x) = \textsc{KNN}(x|D_{train}, B)$, as defined in Section~\ref{sec:approx}, to obtain a bootstrap dataset, $D_{bootstrap}$. We randomly split the bootstrapped dataset $D_{bootstrap} \sim p(D|D_{train})$ into train/test splits to obtain the ``labelled prompt'', $(X_{subtest}, Y_{subtest}, D_{subtrain})$.

\subsubsection{Bootstrapping Small Datasets}

\micp does not run on smaller $N_{train} \le 3000$ datasets. However, bootstrapping can still be used to finetune the model on said datasets to match the downstream dataset distribution. In this case, we sample from $p(D|D_{train})$ by randomly sampling $90\%$ of training samples without replacement to obtain $D_{subtrain}$ and treating the remaining $10\%$ of sample as $X_{subtest}, Y_{subtest}$.

\subsubsection{Finetuning with Adapters}

To prevent overfitting, we only train a small set of new adapter~\citep{houlsby2019parameter,bapna2019simple,hu2021lora,liu2022few} parameters, $\psi$, on $p(D|D_{train})$, without modifying in the pretrained transformer's parameters, $\theta$.\footnote{Adapters are also efficient because only a small number of parameters are updated, $p(\phi)$ is not needed during finetuning, and different downstream datasets share a common pretrained model.} Specifically, we freeze a pretrained \tabpfn transformer, $q_{\theta}(y|x,D)$. Next, for each downstream dataset, $D_{train}$, we add linear adapter layers~\citep{houlsby2019parameter}, $\mathcal{A}_{\psi}^{(D_{train})}$, with parameters $\psi$, to form $q_{\theta,\psi}^{(D_{train})}(y|x,D,q_{\theta},\mathcal{A}_{\psi}^{(D_{train})})$. During finetuning, only $\psi$ is optimized. Intuitively, $q_\theta$ encodes the handcrafted prior, $p(D|\phi)p(\phi)$, and $\mathcal{A}_\psi^{(D_{train})}$ encodes the bootstrapped prior, $p(D|D_{train})$. We illustrate \capfn in Figure~\ref{fig:micp_capfn}.

%% file: sec-experiments.tex
\begin{figure}
  \centering
  \includegraphics[width=0.7\textwidth]{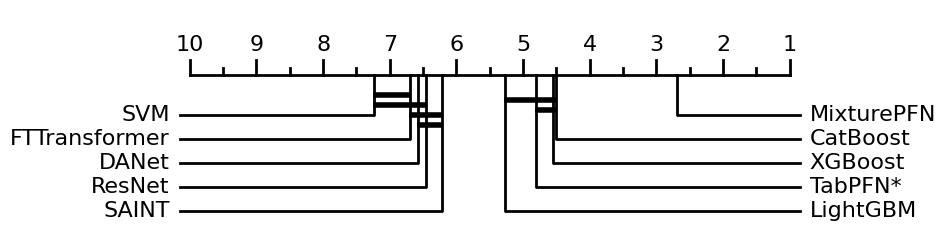}
  \caption{Wilcoxon-Signed Rank Test shows \ours significantly outperforms the Top-10 baselines on the 22 shared datasets. To break ties, we rank algorithms based on their mean log-likelihoods following \tz~\citep{mcelfresh2023neural}. We compute the rank across all 10 cross-validation splits. We report additional critical difference diagrams in the Appendix.}
  \label{fig:critical_diff}
\end{figure}

\begin{table*}
    \centering
        \begin{tabular}{lllll}
            \toprule
            \multirow{ 2}{*}{\textbf{Method}} 
            & Mean $\pm$ Std & Median & Min & Max \\ 
            & Rank$\downarrow$ & Rank$\downarrow$ & Rank$\downarrow$ & Rank$\downarrow$\\
            \midrule
        \ours & \textbf{2.75 $\pm$ 1.94} & \textbf{2} & \textbf{1} & 9 \\ 
\ours (KNNv2) & 4.00 $\pm$ 1.56 & 4 & 2 & \textbf{7} \\ 
CatBoost & 4.70 $\pm$ 3.16 & 6 & \textbf{1} & 9 \\ 
\ours (CaPFN w. Full FT)& 4.85 $\pm$ 2.03 & 4.5 & 2 & 9 \\ 
XGBoost & 4.95 $\pm$ 3.17 & 6 & \textbf{1} & 10 \\ 
\ours (-CaPFN) & 5.10 $\pm$ 1.12 & 5 & 3 & 8 \\ 
\ours (-CaPFN-MICP) = \tabpfnc & 5.30 $\pm$ 1.59 & 5 & 2 & 9 \\ 
\ours (KNNv1) & 7.25 $\pm$ 3.70 & 9 & \textbf{1} & 10 \\ 
MLP-rtdl & 7.25 $\pm$ 2.79 & 8 & \textbf{1} & 10 \\ 
MLP & 8.85 $\pm$ 0.99 & 9 & 7 & 10 \\ 

            \bottomrule
        \end{tabular}
\caption{Ablation table results. \ours (KNNv1) and \ours (KNNv2) replace \micp with a scalable variant of KNN-Prompting. \ours (CaPFN w. Full FT) uses full finetuning instead of adapters. \ours (-CaPFN) and \ours (-CaPFN-MICP) remove each component iteratively, where \micp is replaced by random sampling.}
\label{tab:more}
\end{table*}

\section{Experiment Setup}

We evaluate \ours on the recently proposed \tz benchmark~\citep{mcelfresh2023neural}. \tz is the largest tabular benchmark, with 36 hardest datasets out of 176 tabular classification datasets and 19 baseline algorithms, covering both deep learning and \gbdts. The benchmark covers a diverse range of dataset properties, in number of samples, number of features, and feature distributions. \ours's goal is to (1) improve \tabpfnc~\citep{mcelfresh2023neural}, which randomly samples $B$ training pairs so that \tabpfn~\citep{hollmann2022tabpfn} runs on larger datasets, and (2) outperform both \gbdts~\citep{chen2016xgboost, prokhorenkova2018catboost, ke2017lightgbm} which were found state-of-the-art by \tz, and recent deep learning models~\citep{popov2019neural, gorishniy2021revisiting, arik2021tabnet, hollmann2022tabpfn, yamada2020feature, yoon2020vime, somepalli2021saint, chen2022danets}.

\subsubsection{Evaluation Protocol}
\label{subsubsec:bug}

Since \tz restricts the total runtime to 10 hours, not all algorithms run on the same datasets. To ensure a fair comparison\footnote{We cannot follow the same experimental settings as the October revision of \tz because they are unfair, as mentioned in a recent Github issue~\citep{githubissue}. Further details are in the Appendix.}, we first evaluate \ours and baselines using Condorcet voting, where each dataset ranks the algorithms that successfully ran on it. An algorithm receives a vote whenever it achieves a higher rank than a baseline on each dataset. For each pair of algorithms, the winner received more votes than the loser across all datasets ranking both algorithms. The Condorcet winner is the algorithm that wins all pairwise comparisons.

After determining the Condorcet winner, We further evaluate \ours by ranking it against all baseline algorithms and Top-10 Condorcet algorithms across their shared datasets. We evaluate the performance of each algorithm by its mean rank. Finally, we run a Wilcoxon-Signed Rank Test to check the statistical significance of said mean rank across the Top-10 Condorcet algorithms.

\section{Results}

\subsection{\ours: State-of-the-Art Performance}
\label{sec:main}

As shown in Table~\ref{tab:main}, \ours achieves state-of-the-art performance on \tz across 36 datasets and 19 baseline algorithms. Specifically, \ours is the \textbf{Condorcet winner}, receiving both the most votes and beating all other baselines in pairwise comparisons. Furthermore, \ours achieves the top mean-rank across all subsets of fairly chosen datasets, followed by \gbdts~\citep{chen2016xgboost, prokhorenkova2018catboost}, then \tabpfnc~\citep{mcelfresh2023neural}, then deep learning algorithms~\citep{chen2022danets, gorishniy2021revisiting, somepalli2021saint}. These results corroborate recent findings~\citep{mcelfresh2023neural, grinsztajn2022tree} that most deep learning algorithms fail on general tabular datasets. \ours achieves its state-of-the-art results by scaling \tabpfnc's impressive performance to larger datasets. We provide additional metrics in the Appendix. To understand what dataset regimes each algorithm performs best at, we evaluate \ours's rankings w.r.t. dataset properties.

\textbf{\ours substantially improves the scalability of \tabpfn and \tabpfnc.} As show in Figure~\ref{fig:scaletabpfn}, unlike \tabpfn which encounters memory bottlenecks on datasets with $>3000$ samples, \ours successfully runs on all said datasets. As show in Figure~\ref{fig:scaletabpfn}, \ours substantially improves the performance of \tabpfnc by improving how samples are chosen for the ``prompt'' and training on the downstream dataset.

\textbf{\ours encounters no performance deterioration w.r.t. dataset size.} As shown in Figure~\ref{fig:scaleall}, unlike \tabpfnc, whose performance deteriorates w.r.t  dataset size, \ours's performance compared to the next best baseline is not correlated with dataset size. Hence, \ours is necessary to scale \tabpfnc's impressive performance on to larger datasets.

\textbf{\ours is robust to irregular datasets.} We measure the irregularity of datasets using the standard deviation of the kurtosis of all features. Deep learning algorithms are especially susceptible to irregular datasets with uninformative or heavy-tail features~\citep{grinsztajn2022tree}. As shown in Figure~\ref{fig:scaleprop}, \ours is the Top-1 algorithms on datasets with both high and low kurtosis standard deviation. Because it is finetuned on downstream datasets, \ours is robust to dataset irregularity.

\textbf{\ours works better with fewer features.} As shown in Figure~\ref{fig:scaleprop}, \ours loses against baselines on datasets with a large number of features. \pfn transformers are known to face scalability challenges with number of features~\citep{hollmann2022tabpfn}, due to their handling of column order invariance. We believe better tokenization practices and feature selection can improve feature size scalability, and leave such exploration to future work.

\textbf{\ours's state-of-the-art performance is statistically significant}. Specifically, we run the Wilcoxon Signed-Rank test with $p<0.05$ comparing the Top-10 Condorcet algorithms from Table~\ref{tab:main} across their 22 shared datasets and 10 cross-validation splits. As shown in Figure~\ref{fig:critical_diff}, \ours's state-of-the-art performance is statistically significant.

\subsection{Ablation Study}
\label{sec:abl}

\textbf{Both \micp and \capfn contribute to \ours state-of-the-art results.} We perform ablation studies for \micp and \capfn against common \gbdts and deep learning models across 10 shared datasets. As shown in Table~\ref{tab:more}, each component of the model, \micp and \capfn, contributes to achieving state-of-the-art results. \micp helps by efficiently choosing an effective context for the ``prompt''. \capfn helps by aligning the dataset prior through finetuning the \pfn on \micp's prompting policy. Because overfitting is a well-known issue for deep learning models tackling tabular data~\citep{kadra2021well, grinsztajn2022tree}, adapters are a key component to ensure \capfn aligns the pretrained \tabpfn transformer with the downstream data, without destroying its pretraining prior, $p(D|\phi)p(\phi)$.

\textbf{Under the same GPU resources, KNN-Prompting is much less effective than \micp on tabular datasets}. As described in Section~\ref{subsec:batch}, KNN-Prompting does not support \tabpfn-style batching. To empirically verify that \micp improves KNN-Prompting, we replace \micp in \ours with 2 KNN-Prompting variants: \textbf{\ours(KNNv1)}: Because each prompt contains at most $B+N_{batch}$ tokens, we batch KNN-Prompts by considering $B/N_{batch}$ nearest neighbors instead of $B$-nearest neighbors; \textbf{\ours(KNNv2)}: Because LLM-batching fails due to \tabpfn's ensembling overhead, we remove the ensembling procedure and run KNN-Prompting following LLM-batching~\citep{liu2021makes}, as described in Section~\ref{subsec:batch}. As shown in Table~\ref{tab:more}, both KNN-Prompting variants perform substantially worse than \ours because they compromise an essential component of \pfn, either prompt size or ensembling, for performing efficient inference. The relative rankings suggest prompt size matters more than ensembling.

\section{Limitations}
\label{sec:limits}

As shown in Section~\ref{sec:main}, \ours successfully improves \tabpfn's scalability w.r.t. dataset size to achieve state-of-the-art results on the \tz benchmark. However, we notice that \tabpfnc does not scale  well with feature and label count, relying on ensembling to capture feature and label order invariance. Scaling \tabpfnc to datasets with large number of features and labels can further push ICL performance for tabular learning. While this work covers a large number of diverse datasets, we do not cover huge datasets with billions of samples~\citep{zhu2023xtab, yang2023unitabe}. We leave such investigation to future work.

%% file: sec-conclusion.tex
\section{Conclusion}

In this work, we provide a scalable framework for In-Context Learning (\icl) on tabular datasets. To efficiently construct effective ICL ``prompts'', we propose routing test samples through a Sparse Mixture of In-Context Prompters, \micp. To align the PFN with the inference-time datasets, we propose a novel finetuning policy using bootstrapping, \capfn. Our framework scales \pfns from datasets with 3000 samples to those with much larger number of samples. Our framework, \ours, achieves state-of-the-art performance against 19 deep learning and tree-based baselines across 36 general tabular datasets, establishing a new standard for general tabular learning.

%% file: sec-appendix.tex
\section{Related Work}

\subsection{Tabular Learning Algorithms}
Early tabular learning algorithms are based off decision trees, utilizing boosting, feature encoding, and ensembling~\citep{shwartz2022tabular, borisov2022deep, chen2016xgboost}. Early deep learning algorithms are inspired by decision trees, making them end-to-end learnable~\citep{popov2019neural,katzir2020net,hazimeh2020tree,somepalli2021saint, arik2021tabnet}; however, more thorough benchmarks find decison trees produce more reliable results. Hyperparameter tuning~\citep{kadra2021well} and inductive bias~\citep{grinsztajn2022tree} were identified as key weaknesses in deep learning algorithms. Recent works focus on optimizing transformer models for specialized datasets~\citep{huang2020tabtransformer, gorishniy2021revisiting, gorishniy2022embeddings}, and improving decision tree optimization~\citep{joseph2022gate}, or Bayesian learning~\citep{hollmann2022tabpfn, schafl2022hopular, feuer2023scaling} for small datasets. With the rise of LLMs, pretrained tabular learning models also achieve impressive performance at the cost of billions of training points~\citep{yang2023unitabe,zhu2023xtab}. Of these methods, Prior-Fitted Networks~\citep{feuer2023scaling} were identified as a promising direction in recent benchmarks among deep learning approaches for general tabular learning problems~\citep{mcelfresh2023neural}.

\subsection{Gradient-Boosted Decision Trees}
Gradient-boosted decision trees (\gbdts) remain the preferred algorithm of tabular learning practitioners~\citep{chen2016xgboost, prokhorenkova2018catboost}. Deep learning algorithms~\citep{popov2019neural, gorishniy2021revisiting, somepalli2021saint,arik2021tabnet,yamada2020feature,yoon2020vime} fail on larger benchmarks considering different numbers of samples, numbers of features, feature types, feature distributions, and numbers of labels \citep{mcelfresh2023neural, shwartz2022tabular}.
Because \gbdts utilize boosted gradients and are not rotationally invariant before training,  \gbdt learning algorithms have a better inductive bias than Stochastic Gradient Descent-based (SGD-based) deep learning algorithms~\cite{grinsztajn2022tree}. Thus, \gbdts achieve state-of-the-art performance on medium to large datasets with 3,000 to 1,000,000 samples and competitive performance on smaller datasets~\cite{mcelfresh2023neural, grinsztajn2022tree}. 

\subsection{Prior Fitted Networks}
Prior Fitted Networks (\pfn)~\citep{muller2021transformers} approximates Bayesian inference using a data prior, where a parameterized model is trained to minimize the KL-Divergence between it and the posterior predictive distribution. The proof for \pfns is derived from meta-learning~\citep{gordon2018meta}. \pfns fall under In-Context Learning~\citep{brown2020language}, as the entire training dataset is fed to the model during inference. Hence, \pfns effectively learn the learning algorithm. This approach is particularly effective on tabular data~\citep{mcelfresh2023neural}, where the data prior effectively regularizes model predictions. \tabpfn~\citep{hollmann2022tabpfn} is a \pfn model specific to tabular data that achieves state-of-the-art performance on small datasets. Unlike preliminary works on scaling \tabpfn~\citep{feuer2023scaling}, that benchmark KMeans and Coreset ``promting'', we propose a sparse mixture of KNN prompters, capable of forming scalable batched ``prompts'' and a novel finetuning protocol for the mixture of prompters. Furthermore, our approach not only improves \tabpfn results on large datasets, but achieves state-of-the-art performance across tabular benchmarks~\citep{mcelfresh2023neural}.

\subsection{Mixture of Experts}
Sparsely gated MoE~\citep{shazeer2017outrageously,lewis2021base} shows a significant improvement in model capacity, training time, and accuracy with a gating mechanism. An expert is a sub-network, which better learns to predict similar data points. A gating mechanism, learnable or non-learnable method, decides to route each data point to the most suited experts~\citep{shazeer2017outrageously}. Switch transformers~\citep{fedus2022switch} is a learning to route approach, where it assigns one data point to only one expert, instead of top-k, which reduces computation, while preserving accuracy. However, learnable routing methods require auxiliary load balancing loss function, and further tuning. In~\citep{roller2021hash}, a non-learnable routing method is proposed, which uses a hashing method to assign similar data points to similar experts. They show that this procedure can be better or be competitive with learnable routing MoE methods. However, a hashing method is not necessarily flexible in assigning data points to suitable experts, as it can cause data skewness and choice of hashing function is sensitive to the downstream task. Inspired by these works, we proposed a non-learnable routing mechanism which assigns one data point to one expert, and our routing method finds the most suitable expert based on similarity of data points with a K-Means clustering method. Also, we used experts with shared weights as opposed to general MoE in most of the previous works, as the prompts, not the model, contains  training examples for In-Context Learning. We highlight tackling MoE in the context of prompting is a newly emerging research interest~\citep{anonymous2024mixtureofexperts}, of which we are among the first.

\subsection{Efficient Transformers}
Long sequence inputs have long been studied by the efficient transformer community. Several linear time and space complexity transformers have been proposed~\citep{katharopoulos2020transformers,wang2020linformer,choromanski2020rethinking,qin2022devil}, primarily be SVD decomposing the attention computation. While efficient transformers can help scale the base \pfn model, linear complexity is too expensive for the scale of tabular data in industry. Furthermore, many approaches require finetuning on downstream data~\citep{katharopoulos2020transformers,wang2020linformer} which is nontrivial for \pfn models. Constant time transformers~\citep{zaheer2020big,bulatov2023scaling,chowdhery2023palm} exploit the sequential nature of text data. These methods also do not apply to \pfns for tabular data, as tabular training data is not inherently sequential. Hence, technologies outside of efficient transformers are needed to effectively scale \pfns for tabular learning.

\subsection{In-Context Learning}
In-Context Learning (\icl) prompts transformers with training examples prepended to the desired query~\citep{brown2020language}. Several works attempt to prompt engineer scalable in-context examples for better downstream performance~\citep{hao2022structured}, among which K-Nearest Neighbors emerge as a reliable choice~\citep{liu2021makes, xu2023k}. However, \icl for LLMs consider queries one-at-a-time instead of in batch fashion, prompts are encoded as natural language, and in-context examples can come from a large corpus of natural lanaguage data~\citep{brown2020language}. These properties are not afforded to \pfn-style \icl, where inference is directly run on training set tokens. In addition to scaling ``prompts'', LLMs can also be finetuned on ICL examples to reach better performance~\citep{thoppilan2022lamda,wei2021finetuned}. However, this is due to LLM pretraining objectives misaligning with the ICL task. Such misalignment is not as obvious in the \pfn case, where the transformer is directly trained on the ICL task~\citep{muller2021transformers}. Hence, we propose Sparse Mixture of In-Context Prompters to support batching and our novel bootstrapping algorithm to finetune \pfn models. In the spirit of multi-modal models~\citep{radford2021learning, xu2022introducing}, \tabpfn~\citep{hollmann2022tabpfn} extends \icl techniques from natural language LLMs to tabular data.

\section{Broader Impact Statement}
\label{sec:broad}

This paper presents work whose goal is to advance the field of tabular and in-context learning. Our work reaches a new state-of-the-art tabular classification accuracy, which has broad positive impact for many industries using relational databases and tabular datasets. We hope our impressive results inspire further research into \pfns and ICL for tabular learning. Our work is built on large transformer models, which are known to hallucinate in the natural language domain. While we observe no such behavior on our tabular datasets, we will open source our code, such that practitioners can plug in their own safe transformer models. We feel there are not any other noteworthy negative societal impacts.  

\section{Math Notations}

We summarize our math notations below:
\begin{itemize}
    \item $B$: Number of training samples in prompt
    \item $K$: Number of experts
    \item $D$: A generic dataset
    \item $D_{train}$: The training dataset
    \item $X_{test}$: The test samples
    \item $X_{batch}$: A batch of testing data
    \item $C$: Number of classes
    \item $q_{\theta}$: PFN Model
    \item $A_{\psi}^{D_{train}}$: PFN Model Adapters trained on $D_{train}$
    \item $\theta$: PFN Model Parameters
    \item  $\psi$: PFN Model Adapter Parameters
    \item $\phi$: PFN hypothesis mechanism
    \item $\mathcal{R}$: Router mapping input test points to 1 of $K$ Prompters
    \item $\mathcal{T}_k$: The $k$-th Prompter
    \item     $D_{supp}(x)$: The $B$-Nearest Neighbor training samples to a test point, $x$
    \item $D_{cluster}^{(k)}$: The $k$-th K-Means cluster of training samples
    \item $D_{prompt}^{(k)}$: The $k$-th Prompter’s training samples context
    \item $(X_{batch}, D_{prompt}^{(k)})$: The $k$-th Prompter’s ``prompt''
    \item $NNS(\cdot|\cdot)$: Nearest Neighbor Search Algorithm
    \item $KNN(\cdot|\cdot, \cdot)$: K-Nearest Neighbors Algorithm
    \item $KMeans(\cdot)$: K-Means Algorithm
    \item $Sample(\cdot)$: Random Sampling
    \item $N_{train}$: Full training dataset size
    \item $N_{test}$: Full testing dataset size
    \item $N_{batch}$: Batch size
    \item $\gamma$: Single hyperparameter trading off performance and efficiency.
\end{itemize}

\section{Nonzero Overlap Proof}

\label{sec:proof}

We prove Theorem~\ref{th:overlap1} here.

First we prove $|D_{cluster}^{(k)}| \leq B \implies D_{cluster}^{(k)} \subseteq \textsc{KNN}(x_{center}^{(k)}| D_{train}, B)$:

By KMeans definition, 

$d(x_{center}^{(k)}, x_{train}^{(i)}) < d(x_{center}^{(k)}, x_{train}^{(j)}) \implies (x_{train}^{(j)} \in D_{cluster}^{(k)} \implies x_{train}^{(i)} \in D_{cluster}^{(k)})$,

$\implies \exists \tau_{KMeans}^{(k)} : d(x_{center}^{(k)}, x_{train}^{(i)}) < \tau_{KMeans}^{(k)} \implies x_{train}^{(i)} \in D_{cluster}^{(k)}$

By KNN definition,

$\textsc{KNN}(x| D_{train}, B) = \{x_{train}^{(i)}:d(x_{center}^{(k)}, x_{train}^{(i)}) < \tau_{KNN}(x| D_{train})\}$, 

where $|\{x_{train}^{(i)}:d(x_{center}^{(k)}, x_{train}^{(i)}) < \tau_{KNN}(x| D_{train})\}| = B$

Given $|D_{cluster}^{(k)}| \leq B$,

$\implies |D_{cluster}^{(k)}| \leq |\textsc{KNN}(x_{center}^{(k)}| D_{train}, B)|$

$\implies \textsc{KNN}(x_{center}^{(k)}| D_{train}, B) = D_{cluster}^{(k)} \cup \{ x_{train}^{(i)}:\tau_{KMeans}^{(k)} \leq d(x_{center}^{(k)}, x_{train}^{(i)}) < \tau_{KNN}(x| D_{train})\}$

$\implies D_{cluster}^{(k)} \subseteq \textsc{KNN}(x_{center}^{(k)}| D_{train}, B)$

Next, we prove Theorem~\ref{th:overlap1}:

Given $|D_{cluster}^{(k)}| \leq B$ $\forall k\in[0,...,K-1]$ and $\mathcal{R}^*(x_{train}^{(i)}) = p:x_{train}^{(i)} \in D_{cluster}^{(k)}$ ,

$\implies G(k) = 1$

$\implies x_{train}^{(i)} \in \textsc{KNN}(x_{train}^{(i)}| D_{train}, B) = D_{supp}(x_{train}^{(i)})$

$\implies x_{train}^{(i)} \in D_{cluster}^{(\mathcal{R}^*(x_{train}^{(i)}))} \subseteq \textsc{KNN}(x_{center}^{(\mathcal{R}^*(x_{train}^{(i)}))}| D_{train}, B) = D_{prompt}(\mathcal{R}^*(x_{train}^{(i)}))$

$\implies x_{train}^{(i)} \in D_{prompt}(\mathcal{R}^*(x_{train}^{(i)})) \cap D_{supp}(x_{train}^{(i)})$

$\implies |D_{prompt}(\mathcal{R}^*(x_{train}^{(i)})) \cap D_{supp}(x_{train}^{(i)})| \ge |\{x_{train}^{(i)}\}| = 1$  $\forall i \in [0, ..., N_{train}-1]$ $\forall D_{train}$.

\section{Support Set and KNN Prompting.}

\begin{figure}
  \centering
  \includegraphics[width=0.7\textwidth]{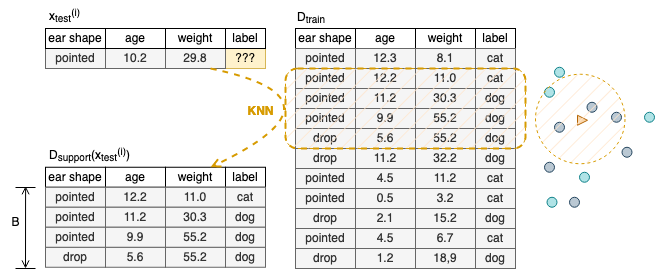}
  \caption{We hypothesize only a subset of the training data, $D_{support}(x_{test}^{(i)})$, is required for effective in-context learning on test sample, $x_{test}^{(i)}$, and this subset is the $B$ nearest training samples in feature space: $D_{support}(x_{test}^{(i)})=\textsc{KNN}(x_{test}^{(i)}|D_{train}, B)$.}
  \label{fig:approx}
\end{figure}

We provide illustrations for the support set as described in Section~\ref{sec:approx} in Figure~\ref{fig:approx}. Note, support sets, as defined in Section~\ref{sec:approx}, are very similar to the KNN-Prompting idea from In-Context Learning (ICL) for Large Language Models (LLMs). We point out one key difference between KNN-Prompting with \tabpfn~\citep{hollmann2022tabpfn} and with ICL on LLMs~\citep{liu2021makes}: LLMs support batching multiple test pairs across multiple ``prompts'' where each ``prompt'' contains the nearest neighbors to 1 test point. In contrast, \tabpfn requires multiple test cases in 1 ``prompt'', which necessitates techniques like \micp to route a batch of test points to the same ``prompt''. As shown in Section~\ref{sec:abl}, \tabpfn's more efficient prompts achieve better performance under same GPU constraints as ICL for LLM batched prompts. We leave application of \micp to ICL and LLMs as future work.

\section{Time and Space Complexity Details}

\textbf{Router and prompter initialization takes $\mathcal{O}(tN_{train}K + (N_{train}+KB)logN_{train})$ time and $\mathcal{O}(N_{train} + KB)$ space complexity and is done once before inference.} For initialization, K-Means with $t$-iterations takes $\mathcal{O}(tN_{train}K)$ time and $\mathcal{O}(K)$ space. To perform efficient nearest neighbor queries, we use the ball-tree algorithm over the training dataset and cluster centers , which takes $\mathcal{O}(N_{train}log(N_{train}))$ time and $\mathcal{O}(N_{train})$ space. Using ball-tree KNN queries, constructing each \icp support set takes $\mathcal{O}(Blog(N_{train}))$ time and $\mathcal{O}(B)$ space.

\textbf{Routing takes $\mathcal{O}(log(K))$ time and $\mathcal{O}(1)$ space complexity, using efficient nearest neighbor search with ball-tree for each test sample.} Router overhead is practically overcome via highly-optimized NNS implementations~\citep{douze2024faiss}, which scale $K$ to the billions.

\textbf{\pfn transformer inference takes $\mathcal{O}(B^2 + BN_{batch})$ time and space complexity, as \micp prompts contain at most $B$ training samples and $N_{batch}=|X_{batch}|$ testing samples.} Note, unlike purely KNN-based prompting, \micp supports batched computation to further amortize \pfn inference cost.

\begin{table*}
\centering
\begin{tabular}{l|l|l}
\toprule
\textbf{Subset} & Considered Algorithms & Considered Datasets \\
\midrule

ALL
& MixturePFN, CatBoost, & ada-agnostic, australian, balance-scale,  \\
& TabPFN*, XGBoost, & colic, credit-approval, elevators, heart-h,  \\
& ResNet, FTTransformer, & jasmine, kc1, lymph, mfeat-fourier, mfeat-zernike,  \\ 
& LightGBM, SAINT, & monks-problems-2, phoneme, profb,  \\
& NODE, MLP-rtdl, & qsar-biodeg, socmob, speeddating, splice, vehicle \\ 
& RandomForest, TabNet,  & \\ 
& MLP, DecisionTree, & \\ 
& LinearModel, STG,  & \\ 
& VIME, KNN, DANet,  & \\
& SVM & \\
\hline
Top-10
& MixturePFN, CatBoost, & ada-agnostic, artificial-characters, australian,  \\ 
& TabPFN*, XGBoost, & balance-scale, colic, credit-approval, elevators,  \\
& ResNet, LightGBM, & gesturephasesegmentationprocessed, heart-h, jasmine,  \\ 
& SAINT, DANet, & kc1, lymph, mfeat-fourier, mfeat-zernike,  \\
& FTTransformer, SVM,  & monks-problems-2, phoneme, profb, qsar-biodeg, \\ 
& & socmob, speeddating, splice, vehicle\\
& & \\
\hline
Top-5
& MixturePFN, CatBoost, & ada-agnostic, artificial-characters, australian,   \\ 
& TabPFN*, XGBoost, & balance-scale, colic, credit-approval, electricity, elevators, \\
& SAINT & albert, gesturephasesegmentationprocessed, heart-h, higgs,   \\
& & jasmine, jungle-chess-2pcs-raw-endgame-complete, kc1, \\
& & lymph, mfeat-fourier, mfeat-zernike, monks-problems-2, \\
& & phoneme, profb, qsar-biodeg, socmob, speeddating, \\
& & splice, vehicle \\

\bottomrule
\end{tabular}
\caption{Datasets and algorithms considered in each Top-K subset. For fair evaluation~\citep{githubissue}, we only consider the shared set of datasets all algorithms run on in the Top-K subsets. By considering different subsets, we evaluate \ours against more or less algorithms and datasets. 20 datasets are shared by all 20 algorithms. 22 datasets are shared by Top-10 algorithms. 25 datasets are shared by Top-5 algorithms. \ours achieves the best mean rank among all Top-K subsets.}
\label{tab:datasetsubset}
\end{table*}

\begin{table*}
\centering
\begin{tabular}{l|llll}
\toprule
\multirow{ 2}{*}{\textbf{Method}} & \multicolumn{4}{c}{\textbf{Condorcet Statistics}} \\
& \#Votes$\uparrow$ & \#Wins$\uparrow$ & \#Ties & \#Losses$\downarrow$ \\
\midrule

MixturePFN & \textbf{464} & \textbf{19} & 0 & \textbf{0} \\
CatBoost & 462 & 18 & 0 & 1 \\
XGBoost & 448 & 16 & 1 & 2 \\
SAINT & 402 & 15 & 0 & 4 \\
ResNet & 367 & 14 & 2 & 3 \\
LightGBM & 360 & 12 & 0 & 7 \\
FTTransformer & 345 & 13 & 1 & 5 \\
TabPFN* & 330 & 11 & 0 & 8 \\
NODE & 304 & 11 & 0 & 8 \\
DANet & 300 & 10 & 1 & 8 \\
RandomForest & 299 & 12 & 1 & 6 \\
MLP-rtdl & 256 & 8 & 0 & 11 \\
SVM & 236 & 6 & 0 & 13 \\
TabNet & 231 & 4 & 0 & 15 \\
MLP & 229 & 7 & 0 & 12 \\
STG & 188 & 5 & 0 & 14 \\
DecisionTree & 155 & 2 & 0 & 17 \\
LinearModel & 144 & 3 & 0 & 16 \\
KNN & 112 & 0 & 0 & 19 \\
VIME & 93 & 1 & 0 & 18 \\

\bottomrule
\end{tabular}
\caption{\ours is the Condorcet winner across 36 datasets against 19 baseline algorithms. We rank algorithms based on their accuracies.}
\label{tab:acc2}
\end{table*}

\begin{table*}
\centering
\begin{tabular}{l|llll}
\toprule
\multirow{ 3}{*}{\textbf{Method}} & \multicolumn{4}{c}{\textbf{All Algorithms (Log Likelihood)}} \\ 
& Mean $\pm$ Std & Median & Min & Max   \\
& Rank$\downarrow$& Rank$\downarrow$& Rank$\downarrow$& Rank$\downarrow$\\
\midrule

MixturePFN & \textbf{2.350 $\pm$ 1.824} & \textbf{1.0} & \textbf{1.0} & \textbf{6.0} \\
TabPFN* & 4.550 $\pm$ 2.747 & 4.5 & 2.0 & 13.0 \\
CatBoost & 4.900 $\pm$ 4.158 & 3.0 & \textbf{1.0} & 14.0 \\
XGBoost & 5.500 $\pm$ 4.621 & 4.0 & \textbf{1.0} & 16.0 \\
SAINT & 8.300 $\pm$ 5.367 & 7.0 & \textbf{1.0} & 19.0 \\
ResNet & 8.400 $\pm$ 3.262 & 8.5 & 3.0 & 15.0 \\
FTTransformer & 8.600 $\pm$ 3.541 & 8.5 & 3.0 & 15.0 \\
DANet & 9.050 $\pm$ 3.369 & 8.5 & 3.0 & 17.0 \\
LightGBM & 9.150 $\pm$ 4.351 & 10.5 & 2.0 & 16.0 \\
MLP-rtdl & 10.800 $\pm$ 5.046 & 10.5 & 2.0 & 20.0 \\
SVM & 11.300 $\pm$ 4.766 & 11.0 & 2.0 & 19.0 \\
RandomForest & 11.600 $\pm$ 4.443 & 12.5 & 4.0 & 18.0 \\
STG & 11.900 $\pm$ 4.549 & 12.5 & 4.0 & 19.0 \\
LinearModel & 12.400 $\pm$ 4.652 & 12.5 & 5.0 & 20.0 \\
NODE & 13.350 $\pm$ 3.410 & 13.5 & 7.0 & 18.0 \\
TabNet & 13.550 $\pm$ 5.296 & 14.0 & 2.0 & 20.0 \\
MLP & 13.700 $\pm$ 3.621 & 14.0 & 4.0 & 19.0 \\
VIME & 15.350 $\pm$ 3.851 & 17.0 & 6.0 & 19.0 \\
DecisionTree & 16.800 $\pm$ 3.881 & 18.5 & 7.0 & 20.0 \\
KNN & 18.450 $\pm$ 1.936 & 19.0 & 14.0 & 20.0 \\

\bottomrule
\end{tabular}
\caption{\ours achieves the top mean rank w.r.t. Log Likelihood across 20 datasets where all algorithms successfully run.}
\label{tab:acc3}
\end{table*}

\begin{table*}
\centering
\begin{tabular}{l|llll}
\toprule
\multirow{ 3}{*}{\textbf{Method}} & \multicolumn{4}{c}{\textbf{All Algorithms (Accuracy)}} \\ 
& Mean $\pm$ Std & Median & Min & Max   \\
& Rank$\downarrow$& Rank$\downarrow$& Rank$\downarrow$& Rank$\downarrow$\\
\midrule
MixturePFN & \textbf{3.950 $\pm$ 3.570} & \textbf{3.0} & \textbf{1.0} & \textbf{13.0} \\
TabPFN* & 5.500 $\pm$ 4.056 & 5.5 & \textbf{1.0} & 15.0 \\
CatBoost & 6.350 $\pm$ 5.313 & 4.0 & \textbf{1.0} & 17.0 \\
XGBoost & 7.100 $\pm$ 5.234 & 5.5 & \textbf{1.0} & 18.0 \\
ResNet & 7.600 $\pm$ 4.017 & 6.5 & \textbf{1.0 }& 16.0 \\
FTTransformer & 7.900 $\pm$ 4.158 & 7.5 & \textbf{1.0} & 17.0 \\
SAINT & 7.950 $\pm$ 5.723 & 5.5 & \textbf{1.0} & 20.0 \\
LightGBM & 8.750 $\pm$ 4.918 & 8.5 & \textbf{1.0} & 17.0 \\
NODE & 8.850 $\pm$ 3.395 & 9.0 & 3.0 & 15.0 \\
MLP-rtdl & 9.150 $\pm$ 5.033 & 9.0 & \textbf{1.0} & 18.0 \\
RandomForest & 9.350 $\pm$ 4.757 & 8.5 & 4.0 & 19.0 \\
DANet & 10.000 $\pm$ 4.405 & 11.0 & 3.0 & 19.0 \\
SVM & 12.250 $\pm$ 5.476 & 14.5 & \textbf{1.0} & 19.0 \\
TabNet & 13.050 $\pm$ 4.780 & 13.5 & 2.0 & 20.0 \\
MLP & 13.100 $\pm$ 4.253 & 14.5 & 5.0 & 18.0 \\
LinearModel & 14.100 $\pm$ 3.846 & 14.0 & 8.0 & 20.0 \\
DecisionTree & 14.250 $\pm$ 4.426 & 15.0 & 3.0 & 20.0 \\
STG & 14.800 $\pm$ 4.423 & 16.0 & 4.0 & 20.0 \\
VIME & 16.950 $\pm$ 2.854 & 18.0 & 10.0 & 20.0 \\
KNN & 17.400 $\pm$ 3.470 & 19.0 & 7.0 & 20.0 \\

\bottomrule
\end{tabular}
\caption{\ours achieves the top mean rank w.r.t. Accuracy across 20 datasets where all algorithms successfully run.}
\label{tab:acc4}
\end{table*}

\begin{table*}
\centering
\begin{tabular}{l|llll}
\toprule
\multirow{ 3}{*}{\textbf{Method}} & \multicolumn{4}{c}{\textbf{Top-10 Algorithms (Log-Likelihood)}} \\ 
& Mean $\pm$ Std & Median & Min & Max   \\
& Rank$\downarrow$& Rank$\downarrow$& Rank$\downarrow$& Rank$\downarrow$\\
\midrule

MixturePFN & \textbf{2.273 $\pm$ 1.710} & \textbf{1.0} & \textbf{1.0} & \textbf{6.0} \\
CatBoost & 3.955 $\pm$ 2.688 & 3.0 & \textbf{1.0} & 10.0 \\
XGBoost & 4.000 $\pm$ 2.663 & 3.0 & \textbf{1.0} & 10.0 \\
TabPFN* & 4.045 $\pm$ 1.965 & 4.0 & 2.0 & 9.0 \\
SAINT & 6.136 $\pm$ 2.473 & 6.5 & \textbf{1.0} & 10.0 \\
LightGBM & 6.409 $\pm$ 2.839 & 7.5 & 2.0 & 10.0 \\
FTTransformer & 6.773 $\pm$ 2.235 & 7.0 & 3.0 & 10.0 \\
ResNet & 6.864 $\pm$ 1.961 & 7.0 & 3.0 & 9.0 \\
DANet & 7.045 $\pm$ 1.988 & 7.0 & 3.0 & 10.0 \\
SVM & 7.500 $\pm$ 2.482 & 8.0 & 2.0 & 10.0 \\

\bottomrule
\end{tabular}
\caption{\ours achieves the top mean rank w.r.t. Log Likelihood across 22 datasets where all Top-10 algorithms successfully run.}
\label{tab:acc5}
\end{table*}

\begin{table*}
\centering
\begin{tabular}{l|llll}
\toprule
\multirow{ 3}{*}{\textbf{Method}} & \multicolumn{4}{c}{\textbf{Top-10 Algorithms (Accuracy)}} \\ 
& Mean $\pm$ Std & Median & Min & Max   \\
& Rank$\downarrow$& Rank$\downarrow$& Rank$\downarrow$& Rank$\downarrow$\\
\midrule

MixturePFN & \textbf{3.000 $\pm$ 2.256} & \textbf{2.0} & \textbf{1.0} & \textbf{9.0} \\
TabPFN* & 4.318 $\pm$ 2.703 & 4.5 & \textbf{1.0} & \textbf{9.0} \\
CatBoost & 4.591 $\pm$ 2.964 & 4.0 & \textbf{1.0} & 10.0 \\
XGBoost & 4.773 $\pm$ 2.859 & 4.0 & \textbf{1.0} & 10.0 \\
ResNet & 5.591 $\pm$ 2.103 & 5.5 & \textbf{1.0} & \textbf{9.0} \\
LightGBM & 5.727 $\pm$ 2.847 & 6.5 & \textbf{1.0} & 10.0 \\
SAINT & 5.727 $\pm$ 2.847 & 5.0 & \textbf{1.0} & 10.0 \\
FTTransformer & 5.864 $\pm$ 2.282 & 6.0 & \textbf{1.0} & \textbf{9.0} \\
DANet & 6.955 $\pm$ 2.184 & 7.5 & 3.0 & 10.0 \\
SVM & 7.773 $\pm$ 2.907 & 9.0 & \textbf{1.0} & 10.0 \\

\bottomrule
\end{tabular}
\caption{\ours achieves the top mean rank w.r.t. Accuracy across 22 datasets where all Top-10 algorithms successfully run.}
\label{tab:acc6}
\end{table*}

\begin{table*}
\centering
\begin{tabular}{l|llll}
\toprule
\multirow{ 3}{*}{\textbf{Method}} & \multicolumn{4}{c}{\textbf{Top-5 Algorithms (Log Likelihood)}} \\ 
& Mean $\pm$ Std & Median & Min & Max   \\
& Rank$\downarrow$& Rank$\downarrow$& Rank$\downarrow$& Rank$\downarrow$\\
\midrule

MixturePFN & \textbf{2.000 $\pm$ 1.166} & \textbf{1.0} & \textbf{1.0} & \textbf{4.0} \\
XGBoost & 2.760 $\pm$ 1.394 & 3.0 & \textbf{1.0} & 5.0 \\
CatBoost & 2.880 $\pm$ 1.243 & 3.0 & \textbf{1.0} & 5.0 \\
TabPFN* & 3.320 $\pm$ 1.085 & 3.0 & 2.0 & 5.0 \\
SAINT & 4.040 $\pm$ 1.311 & 5.0 & \textbf{1.0} & 5.0 \\

\bottomrule
\end{tabular}
\caption{\ours achieves the top mean rank w.r.t. Log Likelihood across 25 datasets where all Top-5 algorithms successfully run.}
\label{tab:acc7}
\end{table*}

\begin{table*}
\centering
\begin{tabular}{l|llll}
\toprule
\multirow{3}{*}{\textbf{Method}} & \multicolumn{4}{c}{\textbf{Top-5 Algorithms (Accuracy)}} \\ 
& Mean $\pm$ Std & Median & Min & Max   \\
& Rank$\downarrow$& Rank$\downarrow$& Rank$\downarrow$& Rank$\downarrow$\\
\midrule

MixturePFN & \textbf{2.360 $\pm$ 1.292} &\textbf{2.0} & \textbf{1.0} & \textbf{5.0 }\\
XGBoost & 2.880 $\pm$ 1.395 & 3.0 & \textbf{1.0} & \textbf{5.0} \\
CatBoost & 2.920 $\pm$ 1.354 & 3.0 & \textbf{1.0} & \textbf{5.0} \\
TabPFN* & 3.000 $\pm$ 1.386 & 3.0 &\textbf{1.0} & \textbf{5.0} \\
SAINT & 3.680 $\pm$ 1.406 & 4.0 & \textbf{1.0} &\textbf{5.0} \\

\bottomrule
\end{tabular}
\caption{\ours achieves the top mean rank w.r.t. Accuracy across 25 datasets where all Top-5 algorithms successfully run.}
\label{tab:acc8}
\end{table*}

\begin{figure}
  \centering
  \includegraphics[width=0.9\textwidth]{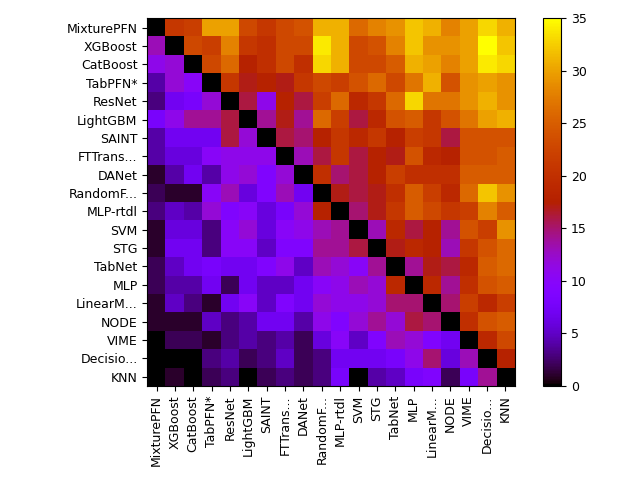}
  \caption{Pairwise comparison matrix for Condorcet voting over the log likelihood metric. Note, \ours is the Condorcet winner.}
  \label{fig:condorcet_acc1}
\end{figure}

\begin{figure}
  \centering
  \includegraphics[width=0.9\textwidth]{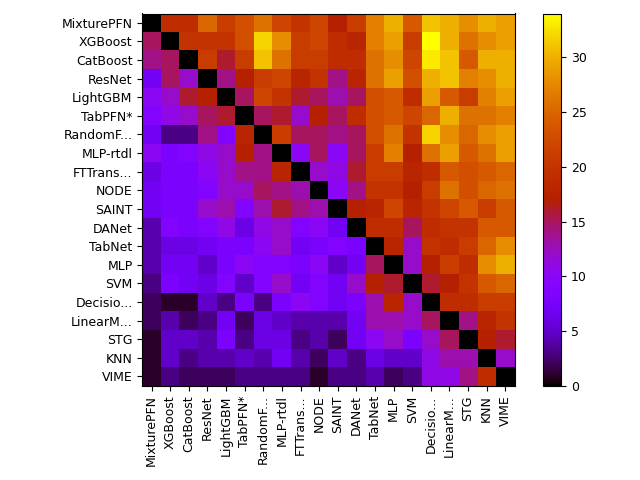}
  \caption{Pairwise comparison matrix for Condorcet voting over the accuracy metric. Note, \ours is the Condorcet winner. Please refer to Section~\ref{sec:additional} for more discussion.}
  \label{fig:condorcet_acc2}
\end{figure}

\begin{figure}
  \centering  
  \subfloat[$N_{train}$ w.r.t. \tabpfnc]{\includegraphics[width=0.275\textwidth]{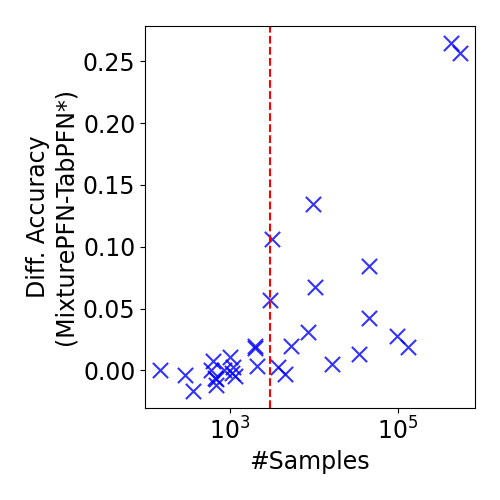}}
  \subfloat[$N_{train}$ w.r.t. baselines]{\includegraphics[width=0.275\textwidth]{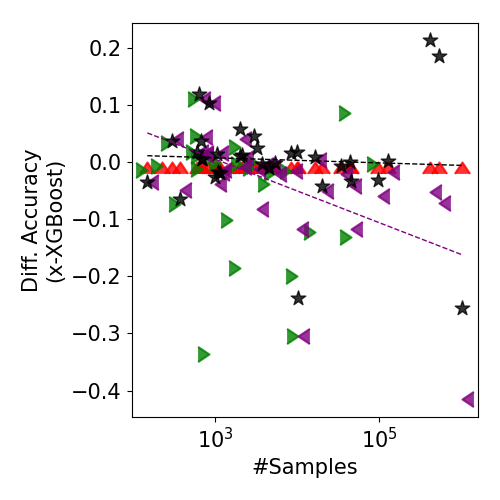}}
  \subfloat[Kurtosis and \#Feat]{\includegraphics[width=0.45\textwidth]{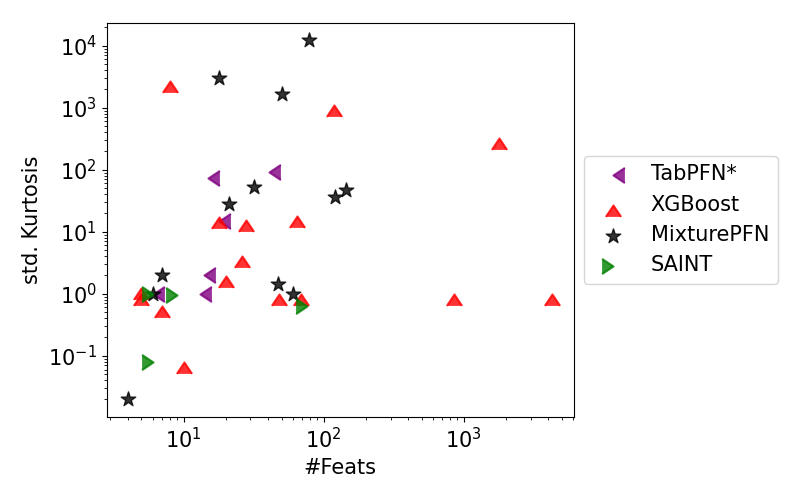}}  
  \caption{We perform the same sensitivity analysis as Figure~\ref{fig:kurtosis} in the main text on the accuracy metric.}
  \label{fig:kurtosis2}
\end{figure}

\begin{figure}
  \centering
  \subfloat[$N_{train}$ w.r.t. baselines]{\includegraphics[width=0.25\textwidth]{figures_new/xgb_samp_nll.png}}
  \subfloat[Kurtosis w.r.t. baselines]{\includegraphics[width=0.25\textwidth]{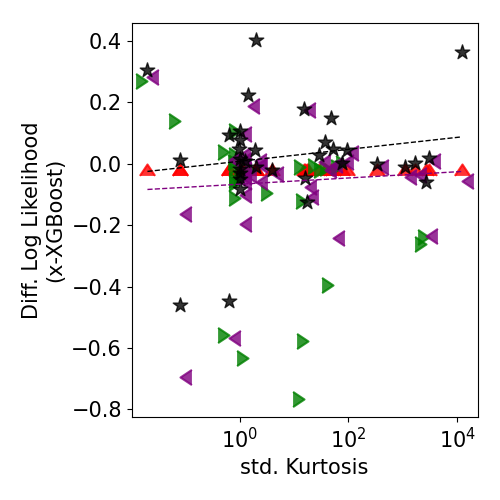}}
  \subfloat[\#Feat w.r.t. baselines]{\includegraphics[width=0.25\textwidth]{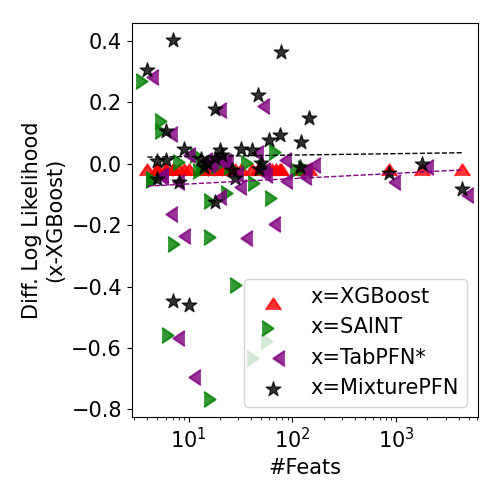}}
  \subfloat[\#Class w.r.t. baselines]{\includegraphics[width=0.25\textwidth]{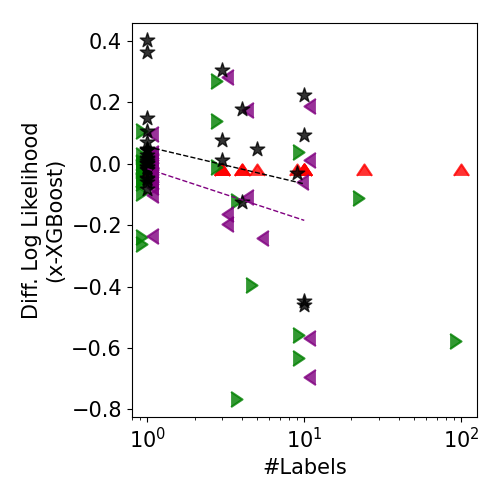}}
  
  \caption{We perform the same sensitivity analysis as Figure~\ref{fig:kurtosis} in the main text but across all dataset properties.}
  \label{fig:kurtosis3}
\end{figure}

\begin{figure}
  \centering
  \subfloat[$N_{train}$ w.r.t. baselines]{\includegraphics[width=0.25\textwidth]{figures_new/xgb_samp_acc.png}}
  \subfloat[Kurtosis w.r.t. baselines]{\includegraphics[width=0.25\textwidth]{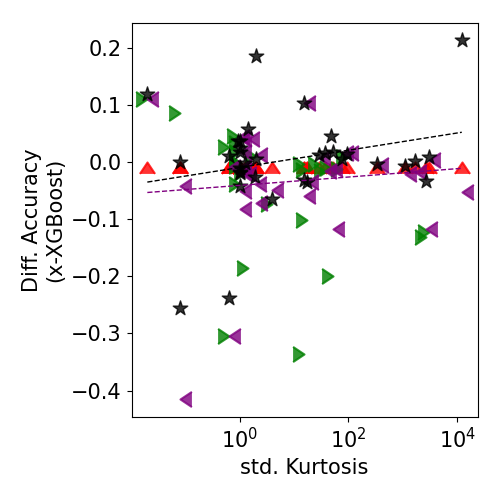}}
  \subfloat[\#Feat w.r.t. baselines]{\includegraphics[width=0.25\textwidth]{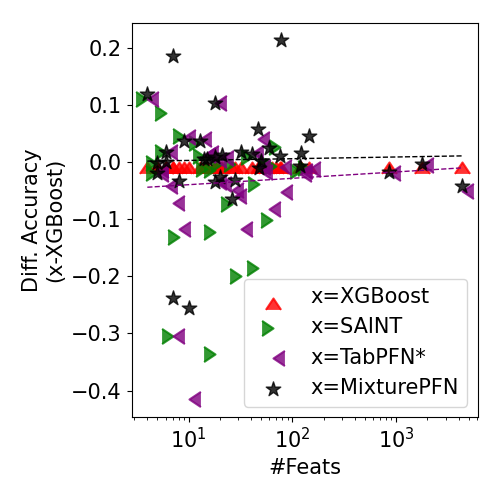}}
  \subfloat[\#Class w.r.t. baselines]{\includegraphics[width=0.25\textwidth]{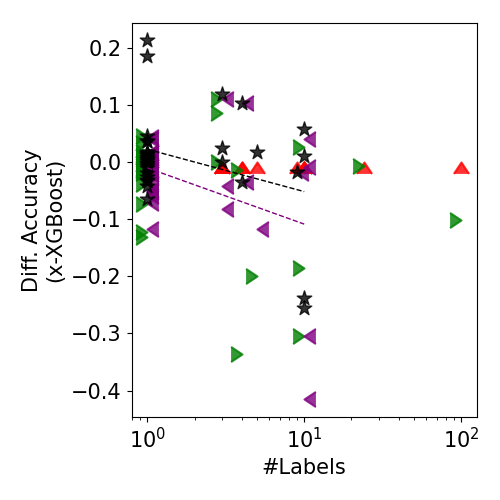}}
  
  \caption{We perform the same sensitivity analysis as Figure~\ref{fig:kurtosis} in the main text but across all dataset properties and on the accuracy metric.}
  \label{fig:kurtosis4}
\end{figure}

\begin{figure}
  \centering
  \includegraphics[width=0.7\textwidth]{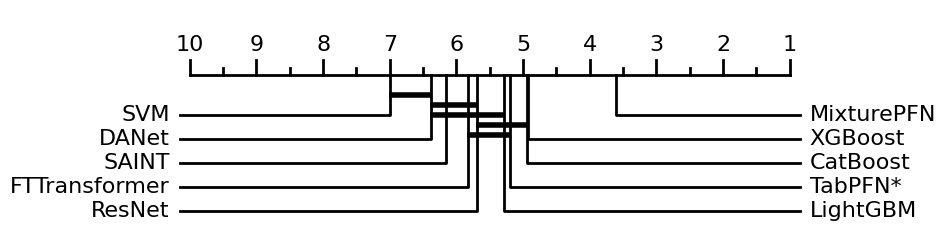}
  \caption{Wilcoxon-Signed Rank Test shows \ours significantly outperforms the Top-10 baselines on the 22 shared datasets, under the accuracy metric. We compute the rank across all 10 cross-validation splits.}
  \label{fig:critical_diff2}
\end{figure}

\begin{figure}
  \centering
  \includegraphics[width=0.7\textwidth]{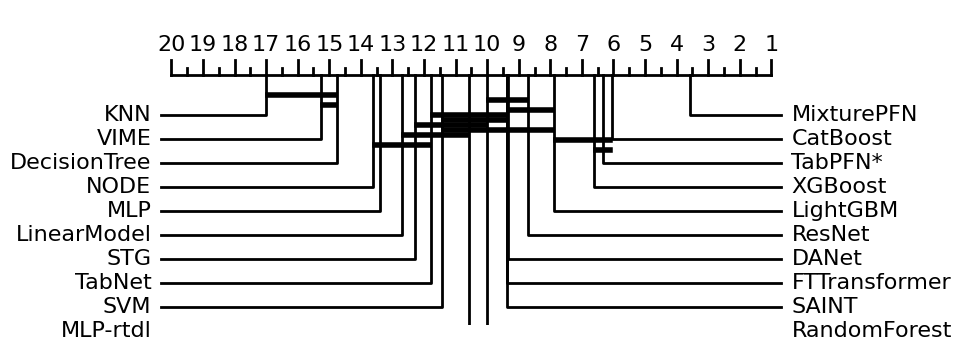}
  \caption{Wilcoxon-Signed Rank Test shows \ours significantly outperforms the all baselines on the 20 shared datasets, under the log likelihood metric. We compute the rank across all 10 cross-validation splits.}
  \label{fig:critical_diff3}
\end{figure}

\begin{figure}
  \centering
  \includegraphics[width=0.7\textwidth]{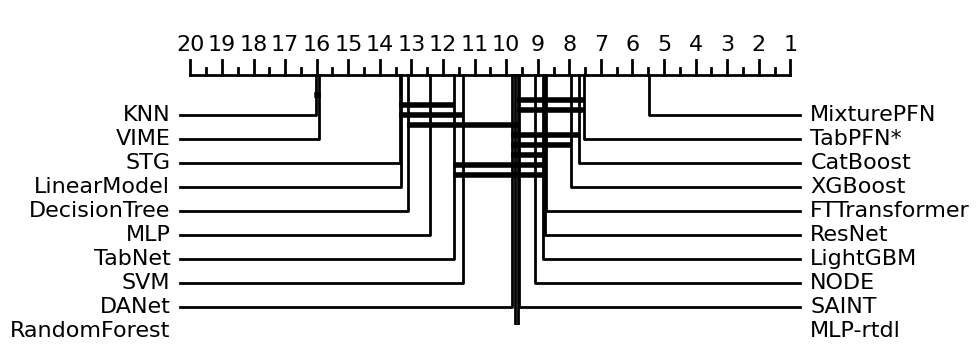}
  \caption{Wilcoxon-Signed Rank Test shows \ours significantly outperforms the all baselines on the 20 shared datasets, under the accuracy metric. We compute the rank across all 10 cross-validation splits.}
  \label{fig:critical_diff4}
\end{figure}

\section{Choosing Baseline Datasets and Algorithms}
\label{sec:datasetbias}

We chose our datasets from the \tz benchmark, which curates 36 of the hardest 176 considered datasets across 19 algorithms. As noted in Section~\ref{subsubsec:bug}, not all algorithms run on all datasets. We note which datasets are shared by all algorithms, Top-10 algorithms, and Top-5 algorithms in Table~\ref{tab:datasetsubset}. We provide the number of datasets each algorithm successfully runs on in Table~\ref{tab:datasetnum}. We provide dataset names and statistics in Table~\ref{tab:data}.

The 2 datasets \ours and \tabpfnc fails on contain >10 classes, which is not currently supported by the pretrained \tabpfn. However, as seen in Figure~\ref{fig:data} and described in Section~\ref{sec:datasets}, we highlight that the 34 datasets \ours and \tabpfnc successfully run on \textbf{cover the full range of dataset properties: number of features, number of samples, and dataset irregularity.} Specifically, our 34 datasets include the ones with the least and most number of samples, the least and most number of features, and the least and most irregularities. Because the focus of this work is to scale \tabpfn to datasets with more number of samples, we leave extending \tabpfn to more number of classes as future work. 

\subsection{Baseline Datasets}
\label{sec:datasets}

\begin{figure}
  \centering
  \includegraphics[width=0.7\textwidth]{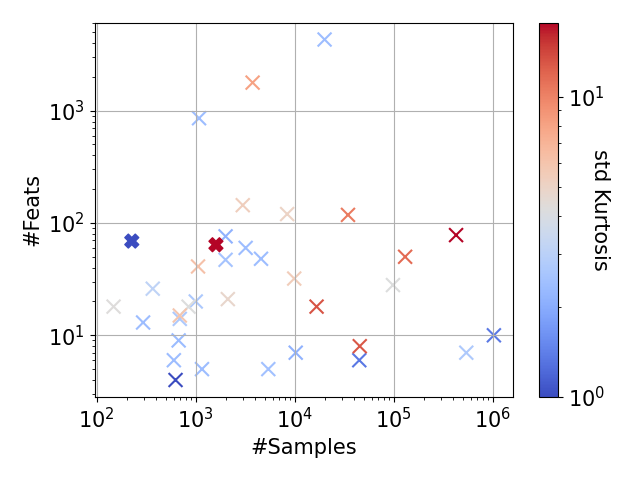}
  \caption{Dataset properties of chosen algorithms from the \tz benchmark. We plot 3 dimensions of the dataset properties of all 36 dataset from \tz. The 2 bold points represent the held-out datasets. As shown, the 34 chosen datasets covers a wide-variety of dataset properties.}
  \label{fig:data}
\end{figure}

We provide dataset statistics in Table~\ref{tab:data}. As shown, our considered datasets cover a wide range of dataset properties in number of features, number of samples, and std. kurtosis. As shown, \ours achieves the best average performance across all datasets. Furthermore, we plot the full range of dataset properties covered, along with the 2 held out datasets from \tz in Figure~\ref{fig:data}, showing that the datasets successfully run on are representative of the benchmark as a whole. Refer to Section~\ref{sec:datasetbias} for more details.

We followed the same experimental setup as TabZilla~\citep{mcelfresh2023neural}, which includes: imputing NaN features to its non-NaN mean and all other preprocessing is handled by each respective baseline. \ours and \tabpfnc follow TabZilla’s PFN preprocessing~\citep{mcelfresh2023neural}: categorical features are encoded as ordinal features, outliers are dropped, features are normalized, results are ensembled across shuffling the feature ordering, and results are ensembled across power-law scaled and unscaled features.

\subsection{Baseline Algorithms}
\label{sec:algorithms}

\subsubsection{Prior-Fitted Network Models (\pfn)}

\tabpfnc is the only \pfn-based baseline, which uses a pretrained 12-layer \tabpfn transformer model, with embeddings size 512, hidden size 1024 in feed-forward layers, and 4-headed attention. \tabpfn is pretrained on a handcrafted dataset prior consisting of randomly generated structural causal models~\citep{hollmann2022tabpfn}. During inference features and labels are randomly shuffled in batch size 32 then ensembled together, following the \tz benchmark~\citep{mcelfresh2023neural}. Our work improves \tabpfn's scalability to different dataset properties, particular in number of training samples.

\subsubsection{Gradient-boosted Decision Tree Models (\gbdt)}

CatBoost~\citep{prokhorenkova2018catboost}, XGBoost~\citep{chen2016xgboost}, and LightGBM~\citep{ke2017lightgbm} are \gbdt models. These models utilize boosting to construct an ensemble of small trees for evaluation. \gbdts are robust to uninformative or heavy tail features and achieve competitive performance over baselines across different dataset properties. Our work argues in-context learning is a potential competitor against \gbdts, as \pfn transformers can potentially learn a better dataset prior than \gbdts.

\subsubsection{Deep Learning Algorithms}

ResNet~\citep{gorishniy2021revisiting}, MLP-rtdl, TabNet~\citep{arik2021tabnet}, MLP, STG~\citep{yamada2020feature}, VIME~\citep{yoon2020vime}, NODE~\citep{popov2019neural}, FTTransformer~\citep{gorishniy2021revisiting}, SVM~\citep{cortes1995support}, DANet~\citep{chen2022danets} and SAINT~\citep{somepalli2021saint} are deep learning-based algorithms. In particular, ResNet is a Convolutional Neural Network designed for tabular learning. MLP-rtdl and MLP are 2 implementations of multilayer-perceptrons. SVM is a support vector machine. TabNet and STG is a neural network architecture that aims to learn \gbdt-like operations in a fully differentiable manner. VIME is a gated neural network that is first traing with self supervision. \ours outperforms deep learning algorithms by learning a prior that better regularizes the learning procedure. NODE is a neural network architecture that aims to imitate \gbdts while being fully differentiable and end2end. DANet is a specialized deep learning architecture for tabular data. FTTransformer is a feature encoding transformer model designed for tabular data. SAINT is a self-supervised transformer designed for tabular data. 

\subsubsection{Simple Algorithms}

RandomForest, DecisionTree, LinearModel, and KNN are all standard machine learning algorithms. We highlight, although \ours is based on evaluating prompts only on KNN neighborhoods, it drastically outperforms KNN. This suggests that In-Context Learning-based models trained on a local neighborhood can outperform both complicated models trained on the entire dataset and simple models that return the average of local neighborhoods. Indeed combining KNN and Large Language Models have been highly successful~\citep{xu2023k,liu2021makes,guu2020retrieval}.

\begin{table*}
\centering
\begin{tabular}{l|l}
\toprule
\textbf{Method} & Number of Datasets Completed On \\ 
\midrule
CatBoost~\citep{prokhorenkova2018catboost} & 35 \\
XGBoost~\citep{chen2016xgboost} & 36 \\
MLP-rtdl~\citep{goodfellow2016deep, gorishniy2021revisiting} & 36 \\
MLP~\citep{goodfellow2016deep, mcelfresh2023neural} & 36 \\
ResNet~\citep{gorishniy2021revisiting} & 35 \\
RandomForest~\citep{liaw2002classification} & 35 \\
DecisionTree~\citep{quinlan1986induction} & 35 \\
MixturePFN & 34 \\
TabPFN*~\citep{hollmann2022tabpfn, mcelfresh2023neural} & 34 \\
LinearModel~\citep{cox1958regression} & 34 \\
TabNet~\citep{arik2021tabnet} & 33 \\
KNN~\citep{cover1967nearest} & 33 \\
LightGBM~\citep{ke2017lightgbm} & 32 \\
VIME~\citep{yoon2020vime} & 32 \\
STG~\citep{yamada2020feature} & 31 \\
NODE~\citep{popov2019neural} & 30 \\
FTTransformer~\citep{gorishniy2021revisiting} & 29 \\
SVM~\citep{cortes1995support} & 29 \\
SAINT~\citep{somepalli2021saint} & 27 \\
DANet~\citep{chen2022danets} & 27 \\
\bottomrule
\end{tabular}
\caption{The number of datasets each algorithm completed on across the entire 36 dataset \tz benchmark. Note, the 2 datasets that \ours and \tabpfnc~\citep{mcelfresh2023neural} does not run on has too many labels, being unsupported by the pretrained \tabpfn~\citep{hollmann2022tabpfn}. However these 2 datasets are not outliers compared to the 34 datasets that are supported. Note, \tabpfn achieves the same results as \tabpfnc, except only running on the 17 datasets with <3,000 features, hence we compare against the more powerful \tabpfnc baseline instead.}
\label{tab:datasetnum}
\end{table*}

\section{Additional Results}
\label{sec:additional}

\subsection{Accuracy Results and Standard Deviations}

We provide the same Condorcet experiments as the main paper but with the accuracy metric in Table~\ref{tab:acc2}. We provide the Condorcet matrix in Figures~\ref{fig:condorcet_acc1} and~\ref{fig:condorcet_acc2}. These results support the same conclusions found in the main paper.

We provide the detailed statistics of the ranking experiments with both log-likelihood and accuracy across all, Top-10, and Top-5 subsets in Tables~\ref{tab:acc3}, \ref{tab:acc4}, \ref{tab:acc5}, \ref{tab:acc6}, \ref{tab:acc7}, \ref{tab:acc8}. These results support the same conclusions found in the main paper.

We provide the Wilcoxon-Signed Rank Test with both the log-likelihood and accuracy metric across the all and Top-10 subsets in Figures~\ref{fig:critical_diff2}, ~\ref{fig:critical_diff3}, and~\ref{fig:critical_diff4}. These results support the same conclusions found in the main paper.

We provide experiments with even lighter hyperparameter tuning, as discussed in Section~\ref{sec:hyperparam}, we call this model, \ours-lite. \ours is tuned over >80\% less configurations than baselines. \ours-lite is tuned only on 2 hyparparameter settings. We provide the main paper results and Condorcet matrices as presented in Table~\ref{tab:acc9} and Figure~\ref{fig:condorcet_acc3}. 

\ours is the Condorcet winner and achieves the top mean rank across all experimental settings, with statistical significance among all and Top-10 subsets. \ours's Log-likelihood results are slightly better, because many algorithms are tied in accuracy across the benchmark. When this occurs, \ours is more confident than baselines when it is correct.

\subsection{Sensitivity Results on More Data}

We provide the same sensitivity analysis conducted in the main paper but with the accuracy metric in Figure~\ref{fig:kurtosis2}. These figures support the same conclusions found in the main paper.

We provide the same sensitivity analysis conducted in the main paper but across the number of features, number of labels, and feature irregularity in Figures~\ref{fig:kurtosis3} and~\ref{fig:kurtosis4}. These figures support the same conclusions found in the main paper.

\begin{table}
\centering
\begin{tabular}{llll}
\toprule
\multirow{ 2}{*}{\textbf{Method}} & \textbf{Accuracy} & \textbf{K} & $\frac{\textbf{Time}}{\textbf{Prompt}}$ \\
& Mean & Mean & Mean \\
\midrule
\tabpfnc & 83.42\% & 1.00 & 1.24s \\ 
T*+\micp ($\gamma=1.0$) & 83.96\% & 2.25 & 0.90s \\ 
T*+\micp ($\gamma=3.0$) & 84.23\% & 5.25 & 1.02s \\ 
T*+\micp ($\gamma=5.0$) & 84.23\% & 8.54 & 0.83s \\ 
\bottomrule
\end{tabular}
\caption{Trade-off of $\gamma$. $T*$+\micp is short for \tabpfnc+\micp. Note as $\gamma$ increases, the accuracy and number of prompters increases, while \tabpfn inference time remains constant. Routing costs are negligible with optimized nearest neighbor search~\citep{douze2024faiss}.}
\label{tab:effeff}
\end{table}

\subsection{Efficiency Effectiveness Trade-Off}
\label{ssubsec:effeff}

\textbf{$\gamma$ effectively trade-offs efficiency and effectiveness.} As mentioned in Section~\ref{sec:trade}, \ours uses a single hyperparameter, $\gamma$, to control the efficiency effectiveness tradeoff. we plot the average accuracy of \tabpfn+\micp across $\gamma=[1.0, 3.0, 5.0]$, across the entire dataset. As shown in Table~\ref{tab:effeff}, as the hyperparameter $\gamma$ increases, \micp's effectiveness is reliably trade-off for efficiency.

\subsection{Timing Analysis}

\begin{figure}
  \centering
  \includegraphics[width=0.5\textwidth]{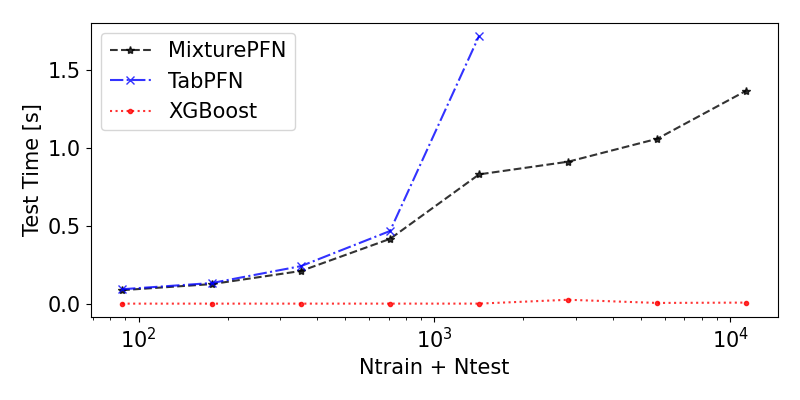}
  \caption{When $N_{train} > 3,000$, \tabpfn runs out of GPU memory. \ours scales much better in runtime than \tabpfn by using \micp to construct bounded size ``prompts''. Inference is slower than XGBoost, due to \tabpfn's transformer's forward pass latency.}
   \label{fig:scal}
\end{figure}

To study runtime, we subsample the electricity dataset into datasets of smaller sizes and then run \tabpfn and \ours. Note, \tabpfn's primary bottleneck is its $\mathcal{O}(N_{train}^2)$ memory bottleneck. While this can be overcome via subsampling, as in the case of \tabpfnc, we study the performance compromises of said approach in Sections~\ref{sec:main} and ~\ref{sec:abl}. As seen in Figure~\ref{fig:scal}, \ours is scalable in both runtime and memory costs compared to \tabpfn.  

\subsection{Detailed Results}

We provide \ours's and \tabpfnc's accuracies across the 10-folds on all datasets in Tables~\ref{tab:allexp} and~\ref{tab:allexp2}.

\section{Implementation}
\label{sec:impl}

We implemented \micp by first preprocessing the train data into separate prompts via KNN, chunking each prompt into batches, then called TabZilla APIs to run the desired PFN model on each batch. We implemented \capfn by bootstrapping our training dataset then running maximum likelihood loss on the bootstrapped datasets. \ours's implementation is built off the official \tz codebase~\citep{githublink}.

\subsection{Optimizing \tabpfn's Implementation}
\tz only obtained \tabpfnc results on 7 out of 34 benchmark datasets~\citep{mcelfresh2023neural}, due to memory constraints. We identified an implementation inefficiency where ``prompts'' are constructed with the entire test dataset, i.e. $(X_{test}, D_{train})$, causing memory overflow. We optimized \tabpfnc's implementation by batching test samples, $(X_{batch}|D_{train}), X_{batch} \subseteq X_{test}$, with batch size 1024, and report results over all 26 datasets.

\begin{table*}
\centering
\begin{tabular}{l|llll}
\toprule
\multirow{ 2}{*}{\textbf{Method}} & \multicolumn{4}{c}{\textbf{Condorcet Statistics}} \\
& \#Votes$\uparrow$ & \#Wins$\uparrow$ & \#Ties & \#Losses$\downarrow$ \\
\midrule

MixturePFN-lite & \textbf{503} & \textbf{19} & 0 & \textbf{0} \\
XGBoost & 502 & 18 & 0 & 1 \\
CatBoost & 479 & 17 & 0 & 2  \\
SAINT & 404 & 16 & 0 & 3\\
TabPFN* & 385 & 13 & 1 & 5\\
LightGBM & 374 & 14 & 1 & 4\\
DANet & 312 & 14 & 0 & 5\\
FTTransformer & 295 & 12 & 0 & 7\\
ResNet & 287 & 10 & 0 & 9\\
SVM & 286 & 11 & 0 & 8\\
STG & 286 & 9 & 0 & 10\\
RandomForest & 248 & 7 & 0 & 12\\
NODE & 244 & 7 & 0 & 12\\
MLP-rtdl & 228 & 5 & 0 & 14\\
TabNet & 210 & 5 & 0 & 14\\
LinearModel & 202 & 3 & 1 & 15\\
MLP & 193 & 5 & 1 & 13\\
VIME & 134 & 2 & 0 & 17\\
DecisionTree & 115 & 1 & 0 & 18\\
KNN & 74 & 0 & 0 & 19\\

\bottomrule
\end{tabular}
\caption{\ours is the Condorcet winner across 36 datasets against 19 baseline algorithms. We rank algorithms based on their log-likelihoods.}
\label{tab:acc9}
\end{table*}

\begin{figure}
  \centering
  \includegraphics[width=0.9\textwidth]{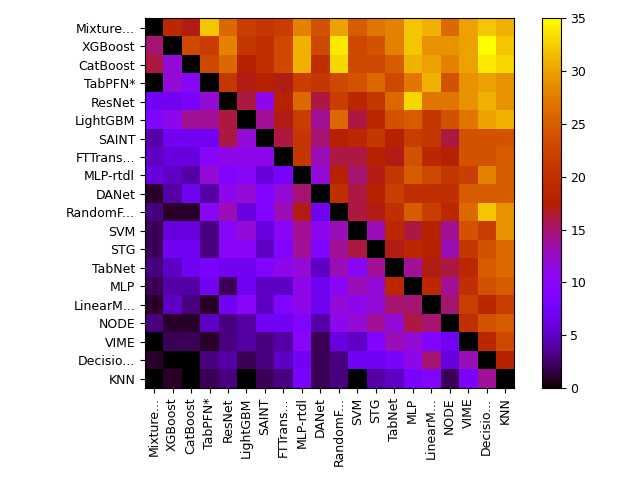}
  \caption{Pairwise comparison matrix for Condorcet voting over the log likelihood metric with lightly tuned \ours. Note, \ours-lite is the Condorcet winner.}
  \label{fig:condorcet_acc3}
\end{figure}

\section{Hyperparameter Setup}
\label{sec:hyperparam}

As \tabpfn transformers can handle up to 3,000 training samples, we set $B=3,000$. We empirically found the minimum number of iterations and batch-size required for loss convergence on the artificial-characters dataset to be 128 iterations and $N_{batch}=64$, which we set for all other datasets. During inference, we use a larger batch size, $N_{batch}=1024$, as gradients no longer need to be stored. We finetuned the model using the Adam optimizer with a learning rate of 0.001. As \tabpfn transformers can handle up to 100 features, for datasets with over 100 features and \tabpfn-based models, we use Maximum Relevance and Minimum Redundancy (mRMR) feature selection~\citep{ding2005minimum} to reduce the number of features to 100. We follow the \tz benchmark, setting $N_{ensemble}=16$, which shuffles features $N_{ensemble}/2$ times for both the original and applies power-law scaled features. \ours's router was implemented using the FAISS~\citep{douze2024faiss} library.

Due to the large variability in datasets in the \tz benchmark, we try 4 hyperparameter settings: (1) $\gamma=5.0$, (2) $\gamma=1.0$, (3) $\gamma=1.0$ but MRMR with 50 features instead of 100 features for feature count scalability, and (4) $\gamma=1.0$ but with Catboost instead of Ordinal encoding for categorical feature scalability~\citep{hollmann2022tabpfn}. Hyperparameters are chosen by picking the setting which maximizes performance on the validation set. Models are evaluated on the test set, which is not seen during hyperparameter tuning. In contrast to all other baselines, which are tuned across 30 hyperparameter settings, \ours performs \textbf{much less} hyperparmeter tuning than baselines. Baseline hyperparameter settings are the same as the \tz~\citep{mcelfresh2023neural} benchmark. Note, even if only the $\gamma$ parameter tuned (i.e. only settings (1) and (2)), \ours is still much better than \tabpfn, as presented in Table~\ref{tab:acc9} and Figure~\ref{fig:condorcet_acc3}. All results were collected over 10-folds following \tz~\citep{mcelfresh2023neural} and OpenML. We tune the hyperparameters by splitting the train set of each fold into training and validation following \tz~\citep{mcelfresh2023neural}. Ablation studies were performed modifying hyperparameter setting (1). Dataset preprocessing details can be found in Appendix~\ref{sec:datasets}.

\section{Hardware}
\label{sec:hardware}

All experiments were conducted on an Nvidia V100 GPU and an AMD EPYC 7402 CPU. Each experiment is given a budget of 10 hours for a single dataset and algorithm.

\begin{table*}
\centering
\begin{tabular}{lllllll}
\toprule
\multirow{ 2}{*}{\textbf{Dataset}} & \multicolumn{4}{c}{\textbf{Dataset Properties}} & \multicolumn{2}{c}{\textbf{Top 2 Algs.}} \\
& \#Samples & \#Feats & \#Lab. & Std. Kurt. & 1st & 2nd \\
\midrule

        lymph & 148 & 18 & 4 & 17.04 & XGBoost & \tabpfnc \\ 
        audiology & 226 & 69 & 24 & None & XGBoost & - \\ 
heart-h & 294 & 13 & 1 & None & MLP-rtdl & \ours \\ 
colic & 368 & 26 & 1 & 4.0 & XGBoost & \ours \\ 
monks-prob... & 601 & 6 & 1 & None & \ours & MLP-rtdl \\ 
balance-scale & 625 & 4 & 3 & 0.02 & \ours & \tabpfnc \\ 
profb & 672 & 9 & 1 & 0.95 & \ours & MLP-rtdl \\ 
Australian & 690 & 14 & 1 & 2.0 & XGBoost & \tabpfnc \\ 
credit-approval & 690 & 15 & 1 & 74.77 & \tabpfnc & \ours \\ 
vehicle & 846 & 18 & 4 & 15.16 & \ours & \tabpfnc \\ 
credit-g & 1000 & 20 & 1 & 1.92 & \ours & \tabpfnc \\ 
qsar-biodeg & 1055 & 41 & 1 & 93.24 & \ours & \tabpfnc \\ 
cnae-9 & 1080 & 856 & 9 & None & MLP-rtdl & MLP \\ 
socmob & 1156 & 5 & 1 & None & XGBoost & \tabpfnc \\ 
100plants & 1599 & 64 & 100 & 17.66 & XGBoost & - \\ 
mfeat-fourier & 2000 & 76 & 10 & 0.64 & \ours & \tabpfnc \\ 
mfeat-zernike & 2000 & 47 & 10 & 1.42 & \ours & \tabpfnc \\ 
kc1 & 2109 & 21 & 1 & 28.34 & \ours & \tabpfnc \\ 
jasmine & 2984 & 144 & 1 & 47.6 & \ours & XGBoost \\ 
splice & 3190 & 60 & 3 & None & \ours & XGBoost \\ 
Bioresponse & 3751 & 1776 & 1 & 328.77 & XGBoost & \ours \\ 
ada-agnostic & 4562 & 48 & 1 & None & XGBoost & \ours \\ 
phoneme & 5404 & 5 & 1 & 1.23 & \ours & XGBoost \\ 
SpeedDating & 8378 & 120 & 1 & 36.43 & \ours & XGBoost \\ 
GesturePhase... & 9873 & 32 & 5 & 52.18 & \ours & XGBoost \\ 
artificial-char... & 10218 & 7 & 10 & 0.63 & XGBoost & \ours \\ 
elevators & 16599 & 18 & 1 & 2986.5 & \ours & \tabpfnc \\ 
guillermo & 20000 & 4296 & 1 & None & XGBoost & \tabpfnc \\ 
nomao & 34465 & 118 & 1 & 1100.34 & XGBoost & \ours \\ 
jungle-chess... & 44819 & 6 & 3 & 0.08 & \ours & XGBoost \\ 
electricity & 45312 & 8 & 1 & 2693.51 & XGBoost & \ours \\ 
higgs & 98050 & 28 & 1 & 15.53 & XGBoost & MLP \\ 
MiniBooNE & 130064 & 50 & 1 & 1686.9 & \ours & XGBoost \\ 
albert & 425240 & 78 & 1 & 12162.65 & \ours & XGBoost \\ 
airlines & 539383 & 7 & 1 & 2.01 & \ours & XGBoost \\ 
poker-hand & 1025009 & 10 & 10 & 0.08 & XGBoost & \ours \\

\bottomrule
\end{tabular}
\caption{Dataset statistics for valid \tz benchmark datasets. Ranks are computed across algorithms that run on all datasets \ours runs on: \ours, \tabpfnc, \xgb, MLP, and MLP-rtdl. Note this list of datasets was originally curated from 197 datasets, to contain only those difficult for all models. We list the top 2 performing algorithms based on log likelihood, following \tz, on each dataset. \ours achieves state-of-the-art performance.}
\label{tab:data}
\end{table*}

\begin{table*}
\centering
\begin{tabular}{lll}
\toprule
Dataset & Model & Mean $\pm$ Std Accuracy $\uparrow$ \\
\midrule
\multirow{2}{*}{australian} & TabPFN* & 0.868$\pm$0.036\\
& MixturePFN & 0.861$\pm$0.023\\
\hline
\multirow{2}{*}{bioresponse} & TabPFN* & 0.791$\pm$0.018\\
& MixturePFN & 0.793$\pm$0.017\\
\hline
\multirow{2}{*}{gesturepha...} & TabPFN* & 0.569$\pm$0.014\\
& MixturePFN & 0.704$\pm$0.011\\
\hline
\multirow{2}{*}{miniboone} & TabPFN* & 0.927$\pm$0.003\\
& MixturePFN & 0.946$\pm$0.003\\
\hline
\multirow{2}{*}{speeddating} & TabPFN* & 0.856$\pm$0.006\\
& MixturePFN & 0.887$\pm$0.011\\
\hline
\multirow{2}{*}{ada-agnostic} & TabPFN* & 0.845$\pm$0.016\\
& MixturePFN & 0.842$\pm$0.013\\
\hline
\multirow{2}{*}{airlines} & TabPFN* & 0.600$\pm$0.003\\
& MixturePFN & 0.857$\pm$0.005\\
\hline
\multirow{2}{*}{albert} & TabPFN* & 0.638$\pm$0.005\\
& MixturePFN & 0.903$\pm$0.003\\
\hline
\multirow{2}{*}{artificial...} & TabPFN* & 0.650$\pm$0.013\\
& MixturePFN & 0.717$\pm$0.008\\
\hline
\multirow{2}{*}{balance-scale} & TabPFN* & 0.989$\pm$0.013\\
& MixturePFN & 0.997$\pm$0.010\\
\hline
\multirow{2}{*}{cnae-9} & TabPFN* & 0.896$\pm$0.029\\
& MixturePFN & 0.899$\pm$0.029\\
\hline
\multirow{2}{*}{colic} & TabPFN* & 0.823$\pm$0.044\\
& MixturePFN & 0.807$\pm$0.067\\
\hline
\multirow{2}{*}{credit-app...} & TabPFN* & 0.884$\pm$0.050\\
& MixturePFN & 0.872$\pm$0.053\\
\hline
\multirow{2}{*}{credit-g} & TabPFN* & 0.729$\pm$0.028\\
& MixturePFN & 0.740$\pm$0.021\\
\hline
\multirow{2}{*}{electricity} & TabPFN* & 0.812$\pm$0.005\\
& MixturePFN & 0.897$\pm$0.003\\
\hline
\multirow{2}{*}{elevators} & TabPFN* & 0.900$\pm$0.006\\
& MixturePFN & 0.905$\pm$0.005\\
\hline
\multirow{2}{*}{guillermo} & TabPFN* & 0.791$\pm$0.013\\
& MixturePFN & 0.799$\pm$0.018\\
\hline
\multirow{2}{*}{heart-h} & TabPFN* & 0.837$\pm$0.044\\
& MixturePFN & 0.834$\pm$0.046\\
\hline
\multirow{2}{*}{higgs} & TabPFN* & 0.665$\pm$0.007\\
& MixturePFN & 0.693$\pm$0.005\\
\hline
\multirow{2}{*}{jasmine} & TabPFN* & 0.804$\pm$0.016\\
& MixturePFN & 0.861$\pm$0.008\\
\hline
\multirow{2}{*}{jungle-che...} & TabPFN* & 0.823$\pm$0.006\\
& MixturePFN & 0.865$\pm$0.004\\
\hline
\multirow{2}{*}{kc1} & TabPFN* & 0.862$\pm$0.011\\
& MixturePFN & 0.866$\pm$0.013\\
\hline
\multirow{2}{*}{lymph} & TabPFN* & 0.810$\pm$0.096\\
& MixturePFN & 0.810$\pm$0.096\\
\hline
\multirow{2}{*}{mfeat-fourier} & TabPFN* & 0.828$\pm$0.025\\
& MixturePFN & 0.847$\pm$0.025\\

\bottomrule
\caption{Mean and Std. Accuracy of \ours and \tabpfnc on all datasets across 10-folds (part 1).}\label{tab:allexp}
\end{tabular}
\end{table*}

\begin{table*}
\centering
\begin{tabular}{lll}
\toprule
Dataset & Model & Mean $\pm$ Std Accuracy $\uparrow$ \\
\midrule

\multirow{2}{*}{mfeat-zernike} & TabPFN* & 0.828$\pm$0.015\\
& MixturePFN & 0.846$\pm$0.024\\
\hline
\multirow{2}{*}{monks-prob...} & TabPFN* & 1.000$\pm$0.000\\
& MixturePFN & 1.000$\pm$0.000\\
\hline
\multirow{2}{*}{nomao} & TabPFN* & 0.953$\pm$0.003\\
& MixturePFN & 0.966$\pm$0.002\\
\hline
\multirow{2}{*}{phoneme} & TabPFN* & 0.883$\pm$0.014\\
& MixturePFN & 0.902$\pm$0.015\\
\hline
\multirow{2}{*}{poker-hand} & TabPFN* & 0.517$\pm$0.011\\
& MixturePFN & 0.677$\pm$0.002\\
\hline
\multirow{2}{*}{profb} & TabPFN* & 0.691$\pm$0.028\\
& MixturePFN & 0.685$\pm$0.024\\
\hline
\multirow{2}{*}{qsar-biodeg} & TabPFN* & 0.885$\pm$0.033\\
& MixturePFN & 0.883$\pm$0.038\\
\hline
\multirow{2}{*}{socmob} & TabPFN* & 0.933$\pm$0.016\\
& MixturePFN & 0.929$\pm$0.017\\
\hline
\multirow{2}{*}{splice} & TabPFN* & 0.876$\pm$0.018\\
& MixturePFN & 0.983$\pm$0.005\\
\hline
\multirow{2}{*}{vehicle} & TabPFN* & 0.847$\pm$0.023\\
& MixturePFN & 0.847$\pm$0.024\\
\bottomrule
\caption{Mean and Std. Accuracy of \ours and \tabpfnc on all datasets across 10-folds (part 2).}\label{tab:allexp2}
\end{tabular}
\end{table*}

\newpage

%% file: neurips_2024.bbl
\begin{thebibliography}{71}
\providecommand{\natexlab}[1]{#1}
\providecommand{\url}[1]{\texttt{#1}}
\expandafter\ifx\csname urlstyle\endcsname\relax
  \providecommand{\doi}[1]{doi: #1}\else
  \providecommand{\doi}{doi: \begingroup \urlstyle{rm}\Url}\fi

\bibitem[abvesa(2024)]{githubissue}
abvesa.
\newblock https://github.com/naszilla/tabzilla/issues/96.
\newblock \emph{Github}, 2024.

\bibitem[Anonymous(2024)]{anonymous2024mixtureofexperts}
Anonymous.
\newblock Mixture-of-experts in prompt optimization, 2024.
\newblock URL \url{https://openreview.net/forum?id=sDmjlpphdB}.

\bibitem[Arik and Pfister(2021)]{arik2021tabnet}
Sercan~{\"O} Arik and Tomas Pfister.
\newblock Tabnet: Attentive interpretable tabular learning.
\newblock In \emph{Proceedings of the AAAI conference on artificial intelligence}, volume~35, pages 6679--6687, 2021.

\bibitem[Bapna et~al.(2019)Bapna, Arivazhagan, and Firat]{bapna2019simple}
Ankur Bapna, Naveen Arivazhagan, and Orhan Firat.
\newblock Simple, scalable adaptation for neural machine translation.
\newblock \emph{arXiv preprint arXiv:1909.08478}, 2019.

\bibitem[Borisov et~al.(2022)Borisov, Leemann, Se{\ss}ler, Haug, Pawelczyk, and Kasneci]{borisov2022deep}
Vadim Borisov, Tobias Leemann, Kathrin Se{\ss}ler, Johannes Haug, Martin Pawelczyk, and Gjergji Kasneci.
\newblock Deep neural networks and tabular data: A survey.
\newblock \emph{IEEE Transactions on Neural Networks and Learning Systems}, 2022.

\bibitem[Bradley et~al.(2000)Bradley, Bennett, and Demiriz]{bradley2000constrained}
Paul~S Bradley, Kristin~P Bennett, and Ayhan Demiriz.
\newblock Constrained k-means clustering.
\newblock \emph{Microsoft Research, Redmond}, 20\penalty0 (0):\penalty0 0, 2000.

\bibitem[Brown et~al.(2020)Brown, Mann, Ryder, Subbiah, Kaplan, Dhariwal, Neelakantan, Shyam, Sastry, Askell, et~al.]{brown2020language}
Tom Brown, Benjamin Mann, Nick Ryder, Melanie Subbiah, Jared~D Kaplan, Prafulla Dhariwal, Arvind Neelakantan, Pranav Shyam, Girish Sastry, Amanda Askell, et~al.
\newblock Language models are few-shot learners.
\newblock \emph{Advances in neural information processing systems}, 33:\penalty0 1877--1901, 2020.

\bibitem[Bulatov et~al.(2023)Bulatov, Kuratov, and Burtsev]{bulatov2023scaling}
Aydar Bulatov, Yuri Kuratov, and Mikhail~S Burtsev.
\newblock Scaling transformer to 1m tokens and beyond with rmt.
\newblock \emph{arXiv preprint arXiv:2304.11062}, 2023.

\bibitem[Chen et~al.(2022)Chen, Liao, Wan, Chen, and Wu]{chen2022danets}
Jintai Chen, Kuanlun Liao, Yao Wan, Danny~Z Chen, and Jian Wu.
\newblock Danets: Deep abstract networks for tabular data classification and regression.
\newblock In \emph{Proceedings of the AAAI Conference on Artificial Intelligence}, volume~36, pages 3930--3938, 2022.

\bibitem[Chen and Guestrin(2016)]{chen2016xgboost}
Tianqi Chen and Carlos Guestrin.
\newblock Xgboost: A scalable tree boosting system.
\newblock In \emph{Proceedings of the 22nd acm sigkdd international conference on knowledge discovery and data mining}, pages 785--794, 2016.

\bibitem[Choromanski et~al.(2020)Choromanski, Likhosherstov, Dohan, Song, Gane, Sarlos, Hawkins, Davis, Mohiuddin, Kaiser, et~al.]{choromanski2020rethinking}
Krzysztof Choromanski, Valerii Likhosherstov, David Dohan, Xingyou Song, Andreea Gane, Tamas Sarlos, Peter Hawkins, Jared Davis, Afroz Mohiuddin, Lukasz Kaiser, et~al.
\newblock Rethinking attention with performers.
\newblock \emph{arXiv preprint arXiv:2009.14794}, 2020.

\bibitem[Chowdhery et~al.(2023)Chowdhery, Narang, Devlin, Bosma, Mishra, Roberts, Barham, Chung, Sutton, Gehrmann, et~al.]{chowdhery2023palm}
Aakanksha Chowdhery, Sharan Narang, Jacob Devlin, Maarten Bosma, Gaurav Mishra, Adam Roberts, Paul Barham, Hyung~Won Chung, Charles Sutton, Sebastian Gehrmann, et~al.
\newblock Palm: Scaling language modeling with pathways.
\newblock \emph{Journal of Machine Learning Research}, 24\penalty0 (240):\penalty0 1--113, 2023.

\bibitem[Cortes and Vapnik(1995)]{cortes1995support}
Corinna Cortes and Vladimir Vapnik.
\newblock Support-vector networks.
\newblock \emph{Machine learning}, 20:\penalty0 273--297, 1995.

\bibitem[Cover and Hart(1967)]{cover1967nearest}
Thomas Cover and Peter Hart.
\newblock Nearest neighbor pattern classification.
\newblock \emph{IEEE transactions on information theory}, 13\penalty0 (1):\penalty0 21--27, 1967.

\bibitem[Cox(1958)]{cox1958regression}
David~R Cox.
\newblock The regression analysis of binary sequences.
\newblock \emph{Journal of the Royal Statistical Society Series B: Statistical Methodology}, 20\penalty0 (2):\penalty0 215--232, 1958.

\bibitem[Ding and Peng(2005)]{ding2005minimum}
Chris Ding and Hanchuan Peng.
\newblock Minimum redundancy feature selection from microarray gene expression data.
\newblock \emph{Journal of bioinformatics and computational biology}, 3\penalty0 (02):\penalty0 185--205, 2005.

\bibitem[Dong et~al.(2022)Dong, Li, Dai, Zheng, Wu, Chang, Sun, Xu, and Sui]{dong2022survey}
Qingxiu Dong, Lei Li, Damai Dai, Ce~Zheng, Zhiyong Wu, Baobao Chang, Xu~Sun, Jingjing Xu, and Zhifang Sui.
\newblock A survey for in-context learning.
\newblock \emph{arXiv preprint arXiv:2301.00234}, 2022.

\bibitem[Douze et~al.(2024)Douze, Guzhva, Deng, Johnson, Szilvasy, Mazaré, Lomeli, Hosseini, and Jégou]{douze2024faiss}
Matthijs Douze, Alexandr Guzhva, Chengqi Deng, Jeff Johnson, Gergely Szilvasy, Pierre-Emmanuel Mazaré, Maria Lomeli, Lucas Hosseini, and Hervé Jégou.
\newblock The faiss library.
\newblock 2024.

\bibitem[Fedus et~al.(2022)Fedus, Zoph, and Shazeer]{fedus2022switch}
William Fedus, Barret Zoph, and Noam Shazeer.
\newblock Switch transformers: Scaling to trillion parameter models with simple and efficient sparsity.
\newblock \emph{The Journal of Machine Learning Research}, 23\penalty0 (1):\penalty0 5232--5270, 2022.

\bibitem[Feuer et~al.(2023)Feuer, Hegde, and Cohen]{feuer2023scaling}
Benjamin Feuer, Chinmay Hegde, and Niv Cohen.
\newblock Scaling tabpfn: Sketching and feature selection for tabular prior-data fitted networks.
\newblock \emph{arXiv preprint arXiv:2311.10609}, 2023.

\bibitem[Goodfellow et~al.(2016)Goodfellow, Bengio, Courville, and Bengio]{goodfellow2016deep}
Ian Goodfellow, Yoshua Bengio, Aaron Courville, and Yoshua Bengio.
\newblock \emph{Deep learning}, volume~1.
\newblock MIT Press, 2016.

\bibitem[Gordon et~al.(2018)Gordon, Bronskill, Bauer, Nowozin, and Turner]{gordon2018meta}
Jonathan Gordon, John Bronskill, Matthias Bauer, Sebastian Nowozin, and Richard~E Turner.
\newblock Meta-learning probabilistic inference for prediction.
\newblock \emph{arXiv preprint arXiv:1805.09921}, 2018.

\bibitem[Gorishniy et~al.(2021)Gorishniy, Rubachev, Khrulkov, and Babenko]{gorishniy2021revisiting}
Yury Gorishniy, Ivan Rubachev, Valentin Khrulkov, and Artem Babenko.
\newblock Revisiting deep learning models for tabular data.
\newblock \emph{Advances in Neural Information Processing Systems}, 34:\penalty0 18932--18943, 2021.

\bibitem[Gorishniy et~al.(2022)Gorishniy, Rubachev, and Babenko]{gorishniy2022embeddings}
Yury Gorishniy, Ivan Rubachev, and Artem Babenko.
\newblock On embeddings for numerical features in tabular deep learning.
\newblock \emph{Advances in Neural Information Processing Systems}, 35:\penalty0 24991--25004, 2022.

\bibitem[Grinsztajn et~al.(2022)Grinsztajn, Oyallon, and Varoquaux]{grinsztajn2022tree}
L{\'e}o Grinsztajn, Edouard Oyallon, and Ga{\"e}l Varoquaux.
\newblock Why do tree-based models still outperform deep learning on typical tabular data?
\newblock \emph{Advances in Neural Information Processing Systems}, 35:\penalty0 507--520, 2022.

\bibitem[Gu et~al.(2023)Gu, Dong, Wei, and Huang]{gu2023pre}
Yuxian Gu, Li~Dong, Furu Wei, and Minlie Huang.
\newblock Pre-training to learn in context.
\newblock \emph{arXiv preprint arXiv:2305.09137}, 2023.

\bibitem[Guu et~al.(2020)Guu, Lee, Tung, Pasupat, and Chang]{guu2020retrieval}
Kelvin Guu, Kenton Lee, Zora Tung, Panupong Pasupat, and Mingwei Chang.
\newblock Retrieval augmented language model pre-training.
\newblock In \emph{International conference on machine learning}, pages 3929--3938. PMLR, 2020.

\bibitem[Hao et~al.(2022)Hao, Sun, Dong, Han, Gu, and Wei]{hao2022structured}
Yaru Hao, Yutao Sun, Li~Dong, Zhixiong Han, Yuxian Gu, and Furu Wei.
\newblock Structured prompting: Scaling in-context learning to 1,000 examples.
\newblock \emph{arXiv preprint arXiv:2212.06713}, 2022.

\bibitem[Hazimeh et~al.(2020)Hazimeh, Ponomareva, Mol, Tan, and Mazumder]{hazimeh2020tree}
Hussein Hazimeh, Natalia Ponomareva, Petros Mol, Zhenyu Tan, and Rahul Mazumder.
\newblock The tree ensemble layer: Differentiability meets conditional computation.
\newblock In \emph{International Conference on Machine Learning}, pages 4138--4148. PMLR, 2020.

\bibitem[Hollmann et~al.(2022)Hollmann, M{\"u}ller, Eggensperger, and Hutter]{hollmann2022tabpfn}
Noah Hollmann, Samuel M{\"u}ller, Katharina Eggensperger, and Frank Hutter.
\newblock Tabpfn: A transformer that solves small tabular classification problems in a second.
\newblock \emph{arXiv preprint arXiv:2207.01848}, 2022.

\bibitem[Houlsby et~al.(2019)Houlsby, Giurgiu, Jastrzebski, Morrone, De~Laroussilhe, Gesmundo, Attariyan, and Gelly]{houlsby2019parameter}
Neil Houlsby, Andrei Giurgiu, Stanislaw Jastrzebski, Bruna Morrone, Quentin De~Laroussilhe, Andrea Gesmundo, Mona Attariyan, and Sylvain Gelly.
\newblock Parameter-efficient transfer learning for nlp.
\newblock In \emph{International Conference on Machine Learning}, pages 2790--2799. PMLR, 2019.

\bibitem[Hu et~al.(2021)Hu, Shen, Wallis, Allen-Zhu, Li, Wang, Wang, and Chen]{hu2021lora}
Edward~J Hu, Yelong Shen, Phillip Wallis, Zeyuan Allen-Zhu, Yuanzhi Li, Shean Wang, Lu~Wang, and Weizhu Chen.
\newblock Lora: Low-rank adaptation of large language models.
\newblock \emph{arXiv preprint arXiv:2106.09685}, 2021.

\bibitem[Huang et~al.(2020)Huang, Khetan, Cvitkovic, and Karnin]{huang2020tabtransformer}
Xin Huang, Ashish Khetan, Milan Cvitkovic, and Zohar Karnin.
\newblock Tabtransformer: Tabular data modeling using contextual embeddings.
\newblock \emph{arXiv preprint arXiv:2012.06678}, 2020.

\bibitem[Joseph and Raj(2022)]{joseph2022gate}
Manu Joseph and Harsh Raj.
\newblock Gate: Gated additive tree ensemble for tabular classification and regression.
\newblock \emph{arXiv preprint arXiv:2207.08548}, 2022.

\bibitem[Kadra et~al.(2021)Kadra, Lindauer, Hutter, and Grabocka]{kadra2021well}
Arlind Kadra, Marius Lindauer, Frank Hutter, and Josif Grabocka.
\newblock Well-tuned simple nets excel on tabular datasets.
\newblock \emph{Advances in neural information processing systems}, 34:\penalty0 23928--23941, 2021.

\bibitem[Katharopoulos et~al.(2020)Katharopoulos, Vyas, Pappas, and Fleuret]{katharopoulos2020transformers}
Angelos Katharopoulos, Apoorv Vyas, Nikolaos Pappas, and Fran{\c{c}}ois Fleuret.
\newblock Transformers are rnns: Fast autoregressive transformers with linear attention.
\newblock In \emph{International conference on machine learning}, pages 5156--5165. PMLR, 2020.

\bibitem[Katzir et~al.(2020)Katzir, Elidan, and El-Yaniv]{katzir2020net}
Liran Katzir, Gal Elidan, and Ran El-Yaniv.
\newblock Net-dnf: Effective deep modeling of tabular data.
\newblock In \emph{International conference on learning representations}, 2020.

\bibitem[Ke et~al.(2017)Ke, Meng, Finley, Wang, Chen, Ma, Ye, and Liu]{ke2017lightgbm}
Guolin Ke, Qi~Meng, Thomas Finley, Taifeng Wang, Wei Chen, Weidong Ma, Qiwei Ye, and Tie-Yan Liu.
\newblock Lightgbm: A highly efficient gradient boosting decision tree.
\newblock \emph{Advances in neural information processing systems}, 30, 2017.

\bibitem[Khandelwal et~al.(2019)Khandelwal, Levy, Jurafsky, Zettlemoyer, and Lewis]{khandelwal2019generalization}
Urvashi Khandelwal, Omer Levy, Dan Jurafsky, Luke Zettlemoyer, and Mike Lewis.
\newblock Generalization through memorization: Nearest neighbor language models.
\newblock \emph{arXiv preprint arXiv:1911.00172}, 2019.

\bibitem[Lewis et~al.(2021)Lewis, Bhosale, Dettmers, Goyal, and Zettlemoyer]{lewis2021base}
Mike Lewis, Shruti Bhosale, Tim Dettmers, Naman Goyal, and Luke Zettlemoyer.
\newblock Base layers: Simplifying training of large, sparse models.
\newblock In \emph{International Conference on Machine Learning}, pages 6265--6274. PMLR, 2021.

\bibitem[Liaw et~al.(2002)Liaw, Wiener, et~al.]{liaw2002classification}
Andy Liaw, Matthew Wiener, et~al.
\newblock Classification and regression by randomforest.
\newblock \emph{R news}, 2\penalty0 (3):\penalty0 18--22, 2002.

\bibitem[Liu et~al.(2022)Liu, Tam, Muqeeth, Mohta, Huang, Bansal, and Raffel]{liu2022few}
Haokun Liu, Derek Tam, Mohammed Muqeeth, Jay Mohta, Tenghao Huang, Mohit Bansal, and Colin~A Raffel.
\newblock Few-shot parameter-efficient fine-tuning is better and cheaper than in-context learning.
\newblock \emph{Advances in Neural Information Processing Systems}, 35:\penalty0 1950--1965, 2022.

\bibitem[Liu et~al.(2021)Liu, Shen, Zhang, Dolan, Carin, and Chen]{liu2021makes}
Jiachang Liu, Dinghan Shen, Yizhe Zhang, Bill Dolan, Lawrence Carin, and Weizhu Chen.
\newblock What makes good in-context examples for gpt-$3 $?
\newblock \emph{arXiv preprint arXiv:2101.06804}, 2021.

\bibitem[McElfresh et~al.(2023)McElfresh, Khandagale, Valverde, Ramakrishnan, Goldblum, White, et~al.]{mcelfresh2023neural}
Duncan McElfresh, Sujay Khandagale, Jonathan Valverde, Ganesh Ramakrishnan, Micah Goldblum, Colin White, et~al.
\newblock When do neural nets outperform boosted trees on tabular data?
\newblock \emph{arXiv preprint arXiv:2305.02997}, 2023.

\bibitem[M{\"u}ller et~al.(2021)M{\"u}ller, Hollmann, Arango, Grabocka, and Hutter]{muller2021transformers}
Samuel M{\"u}ller, Noah Hollmann, Sebastian~Pineda Arango, Josif Grabocka, and Frank Hutter.
\newblock Transformers can do bayesian inference.
\newblock \emph{arXiv preprint arXiv:2112.10510}, 2021.

\bibitem[naszilla(2024)]{githublink}
naszilla.
\newblock https://github.com/naszilla/tabzilla.
\newblock \emph{Github}, 2024.

\bibitem[Pearl(2009)]{pearl2009causality}
Judea Pearl.
\newblock \emph{Causality}.
\newblock Cambridge university press, 2009.

\bibitem[Peters et~al.(2017)Peters, Janzing, and Sch{\"o}lkopf]{peters2017elements}
Jonas Peters, Dominik Janzing, and Bernhard Sch{\"o}lkopf.
\newblock \emph{Elements of causal inference: foundations and learning algorithms}.
\newblock The MIT Press, 2017.

\bibitem[Popov et~al.(2019)Popov, Morozov, and Babenko]{popov2019neural}
Sergei Popov, Stanislav Morozov, and Artem Babenko.
\newblock Neural oblivious decision ensembles for deep learning on tabular data.
\newblock \emph{arXiv preprint arXiv:1909.06312}, 2019.

\bibitem[Prokhorenkova et~al.(2018)Prokhorenkova, Gusev, Vorobev, Dorogush, and Gulin]{prokhorenkova2018catboost}
Liudmila Prokhorenkova, Gleb Gusev, Aleksandr Vorobev, Anna~Veronika Dorogush, and Andrey Gulin.
\newblock Catboost: unbiased boosting with categorical features.
\newblock \emph{Advances in neural information processing systems}, 31, 2018.

\bibitem[Qin et~al.(2022)Qin, Han, Sun, Li, Kong, Barnes, and Zhong]{qin2022devil}
Zhen Qin, Xiaodong Han, Weixuan Sun, Dongxu Li, Lingpeng Kong, Nick Barnes, and Yiran Zhong.
\newblock The devil in linear transformer.
\newblock \emph{arXiv preprint arXiv:2210.10340}, 2022.

\bibitem[Quinlan(1986)]{quinlan1986induction}
J.~Ross Quinlan.
\newblock Induction of decision trees.
\newblock \emph{Machine learning}, 1:\penalty0 81--106, 1986.

\bibitem[Radford et~al.(2021)Radford, Kim, Hallacy, Ramesh, Goh, Agarwal, Sastry, Askell, Mishkin, Clark, et~al.]{radford2021learning}
Alec Radford, Jong~Wook Kim, Chris Hallacy, Aditya Ramesh, Gabriel Goh, Sandhini Agarwal, Girish Sastry, Amanda Askell, Pamela Mishkin, Jack Clark, et~al.
\newblock Learning transferable visual models from natural language supervision.
\newblock In \emph{International conference on machine learning}, pages 8748--8763. PMLR, 2021.

\bibitem[Roller et~al.(2021)Roller, Sukhbaatar, Weston, et~al.]{roller2021hash}
Stephen Roller, Sainbayar Sukhbaatar, Jason Weston, et~al.
\newblock Hash layers for large sparse models.
\newblock \emph{Advances in Neural Information Processing Systems}, 34:\penalty0 17555--17566, 2021.

\bibitem[Sch{\"a}fl et~al.(2022)Sch{\"a}fl, Gruber, Bitto-Nemling, and Hochreiter]{schafl2022hopular}
Bernhard Sch{\"a}fl, Lukas Gruber, Angela Bitto-Nemling, and Sepp Hochreiter.
\newblock Hopular: Modern hopfield networks for tabular data.
\newblock \emph{arXiv preprint arXiv:2206.00664}, 2022.

\bibitem[Shazeer et~al.(2017)Shazeer, Mirhoseini, Maziarz, Davis, Le, Hinton, and Dean]{shazeer2017outrageously}
Noam Shazeer, Azalia Mirhoseini, Krzysztof Maziarz, Andy Davis, Quoc Le, Geoffrey Hinton, and Jeff Dean.
\newblock Outrageously large neural networks: The sparsely-gated mixture-of-experts layer.
\newblock \emph{arXiv preprint arXiv:1701.06538}, 2017.

\bibitem[Shwartz-Ziv and Armon(2022)]{shwartz2022tabular}
Ravid Shwartz-Ziv and Amitai Armon.
\newblock Tabular data: Deep learning is not all you need.
\newblock \emph{Information Fusion}, 81:\penalty0 84--90, 2022.

\bibitem[Somepalli et~al.(2021)Somepalli, Goldblum, Schwarzschild, Bruss, and Goldstein]{somepalli2021saint}
Gowthami Somepalli, Micah Goldblum, Avi Schwarzschild, C~Bayan Bruss, and Tom Goldstein.
\newblock Saint: Improved neural networks for tabular data via row attention and contrastive pre-training.
\newblock \emph{arXiv preprint arXiv:2106.01342}, 2021.

\bibitem[Thoppilan et~al.(2022)Thoppilan, De~Freitas, Hall, Shazeer, Kulshreshtha, Cheng, Jin, Bos, Baker, Du, et~al.]{thoppilan2022lamda}
Romal Thoppilan, Daniel De~Freitas, Jamie Hall, Noam Shazeer, Apoorv Kulshreshtha, Heng-Tze Cheng, Alicia Jin, Taylor Bos, Leslie Baker, Yu~Du, et~al.
\newblock Lamda: Language models for dialog applications.
\newblock \emph{arXiv preprint arXiv:2201.08239}, 2022.

\bibitem[Vaswani et~al.(2017)Vaswani, Shazeer, Parmar, Uszkoreit, Jones, Gomez, Kaiser, and Polosukhin]{vaswani2017attention}
Ashish Vaswani, Noam Shazeer, Niki Parmar, Jakob Uszkoreit, Llion Jones, Aidan~N Gomez, {\L}ukasz Kaiser, and Illia Polosukhin.
\newblock Attention is all you need.
\newblock \emph{Advances in neural information processing systems}, 30, 2017.

\bibitem[Von~Oswald et~al.(2023)Von~Oswald, Niklasson, Randazzo, Sacramento, Mordvintsev, Zhmoginov, and Vladymyrov]{von2023transformers}
Johannes Von~Oswald, Eyvind Niklasson, Ettore Randazzo, Jo{\~a}o Sacramento, Alexander Mordvintsev, Andrey Zhmoginov, and Max Vladymyrov.
\newblock Transformers learn in-context by gradient descent.
\newblock In \emph{International Conference on Machine Learning}, pages 35151--35174. PMLR, 2023.

\bibitem[Wang et~al.(2020)Wang, Li, Khabsa, Fang, and Ma]{wang2020linformer}
Sinong Wang, Belinda~Z Li, Madian Khabsa, Han Fang, and Hao Ma.
\newblock Linformer: Self-attention with linear complexity.
\newblock \emph{arXiv preprint arXiv:2006.04768}, 2020.

\bibitem[Wei et~al.(2021)Wei, Bosma, Zhao, Guu, Yu, Lester, Du, Dai, and Le]{wei2021finetuned}
Jason Wei, Maarten Bosma, Vincent~Y Zhao, Kelvin Guu, Adams~Wei Yu, Brian Lester, Nan Du, Andrew~M Dai, and Quoc~V Le.
\newblock Finetuned language models are zero-shot learners.
\newblock \emph{arXiv preprint arXiv:2109.01652}, 2021.

\bibitem[Xu et~al.(2023)Xu, Wang, Mao, Lyu, She, and Zhang]{xu2023k}
Benfeng Xu, Quan Wang, Zhendong Mao, Yajuan Lyu, Qiaoqiao She, and Yongdong Zhang.
\newblock $ k $ nn prompting: Beyond-context learning with calibration-free nearest neighbor inference.
\newblock \emph{arXiv preprint arXiv:2303.13824}, 2023.

\bibitem[Xu et~al.(2022)Xu, Dong, Wang, Kim, Lin, Shrivastava, Li, Tseng, Baevski, Lin, et~al.]{xu2022introducing}
Derek Xu, Shuyan Dong, Changhan Wang, Suyoun Kim, Zhaojiang Lin, Akshat Shrivastava, Shang-Wen Li, Liang-Hsuan Tseng, Alexei Baevski, Guan-Ting Lin, et~al.
\newblock Introducing semantics into speech encoders.
\newblock \emph{arXiv preprint arXiv:2211.08402}, 2022.

\bibitem[Xu et~al.(2024)Xu, Liu, Pasupat, Kazemi, et~al.]{xu2024context}
Xin Xu, Yue Liu, Panupong Pasupat, Mehran Kazemi, et~al.
\newblock In-context learning with retrieved demonstrations for language models: A survey.
\newblock \emph{arXiv preprint arXiv:2401.11624}, 2024.

\bibitem[Yamada et~al.(2020)Yamada, Lindenbaum, Negahban, and Kluger]{yamada2020feature}
Yutaro Yamada, Ofir Lindenbaum, Sahand Negahban, and Yuval Kluger.
\newblock Feature selection using stochastic gates.
\newblock In \emph{International Conference on Machine Learning}, pages 10648--10659. PMLR, 2020.

\bibitem[Yang et~al.(2023)Yang, Wang, Liu, Wu, and Liu]{yang2023unitabe}
Yazheng Yang, Yuqi Wang, Guang Liu, Ledell Wu, and Qi~Liu.
\newblock Unitabe: Pretraining a unified tabular encoder for heterogeneous tabular data.
\newblock \emph{arXiv preprint arXiv:2307.09249}, 2023.

\bibitem[Yoon et~al.(2020)Yoon, Zhang, Jordon, and van~der Schaar]{yoon2020vime}
Jinsung Yoon, Yao Zhang, James Jordon, and Mihaela van~der Schaar.
\newblock Vime: Extending the success of self-and semi-supervised learning to tabular domain.
\newblock \emph{Advances in Neural Information Processing Systems}, 33:\penalty0 11033--11043, 2020.

\bibitem[Zaheer et~al.(2020)Zaheer, Guruganesh, Dubey, Ainslie, Alberti, Ontanon, Pham, Ravula, Wang, Yang, et~al.]{zaheer2020big}
Manzil Zaheer, Guru Guruganesh, Kumar~Avinava Dubey, Joshua Ainslie, Chris Alberti, Santiago Ontanon, Philip Pham, Anirudh Ravula, Qifan Wang, Li~Yang, et~al.
\newblock Big bird: Transformers for longer sequences.
\newblock \emph{Advances in neural information processing systems}, 33:\penalty0 17283--17297, 2020.

\bibitem[Zhu et~al.(2023)Zhu, Shi, Erickson, Li, Karypis, and Shoaran]{zhu2023xtab}
Bingzhao Zhu, Xingjian Shi, Nick Erickson, Mu~Li, George Karypis, and Mahsa Shoaran.
\newblock Xtab: Cross-table pretraining for tabular transformers.
\newblock \emph{arXiv preprint arXiv:2305.06090}, 2023.

\end{thebibliography}
